\documentclass[10pt,twocolumn,letterpaper]{article}

\usepackage{wacv}              % To produce the CAMERA-READY version
\definecolor{wacvblue}{rgb}{0.21,0.49,0.74}
\definecolor{lightgray}{rgb}{0.83, 0.83, 0.83}
\usepackage[pagebackref,breaklinks,colorlinks,allcolors=wacvblue]{hyperref}
\usepackage{pifont}     % For \ding symbols
\usepackage{amsfonts}
\usepackage{bbm}
\usepackage{tikz}
\usepackage[table]{xcolor}
\usepackage{algorithm2e}
\usepackage{listings}
\usepackage{xcolor} % Necesario para definir colores

\lstdefinelanguage{json}{
    keywords={true, false, null},
    keywordstyle=\color{blue}\bfseries,
    stringstyle=\color{red!80!black},
    numberstyle=\color{purple},
    morestring=[b]",
    morecomment=[l]{//}, % JSON no tiene comentarios, pero se añade por si acaso
    commentstyle=\color{gray}\itshape,
    literate=
        *{:}{{{\color{black}:}}}1
        {,}{{{\color{black},}}}1
        {\{}{{{\color{blue}\{}}}1
        {\}}{{{\color{blue}\}}}}1
        {[}{{{\color{blue}[}}}1
        {]}{{{\color{blue}]}}}1,
    alsoletter={-},
}

\newcommand*\circled[1]{\tikz[baseline=(char.base)]{
            \node[shape=circle,draw,inner sep=0.5pt] (char) {#1};}}

\newcommand{\cmark}{\textcolor{green}{\checkmark}} % Green checkmark
\newcommand{\xmark}{\textcolor{red}{\ding{55}}}    % Red cross

\def\wacvPaperID{545} % *** Enter the WACV Paper ID here
\def\confName{WACV}
\def\confYear{2026}

\title{FALCONEye: Finding Answers and Localizing Content\\ in ONE-hour-long videos with multi-modal LLMs}

\author{Carlos Plou \qquad Cesar Borja \qquad Ruben Martinez-Cantin \qquad Ana C. Murillo \\
DIIS-I3A, University of Zaragoza, Spain \\
{\tt\small \{c.plou, cborja, rmcantin, acm\}@unizar.es}
}

\begin{document}
\maketitle

\begin{figure*}
    \centering
    \includegraphics[width=0.8\linewidth]{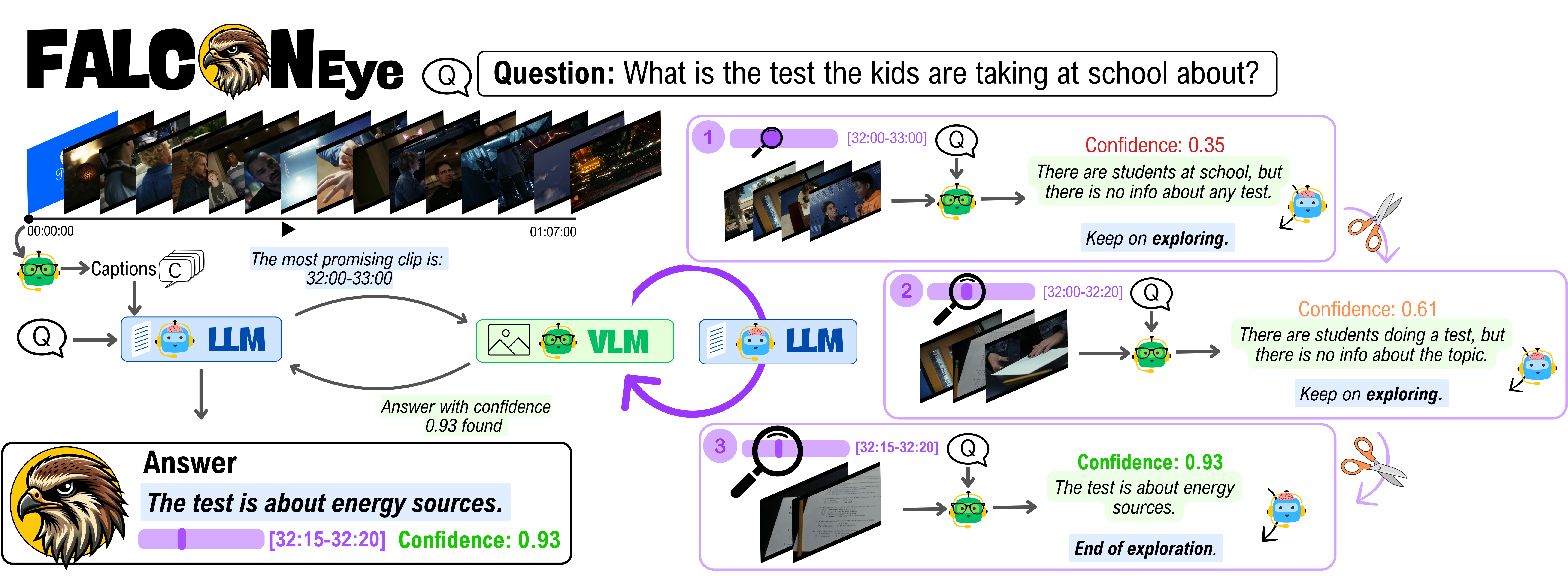}
    \caption{\textbf{Overview of our meta-architecture FALCONEye designed to Finding Answers and Localizing Content in ONE-hour-long videos with multi-modal LLMs}. A VLM pre-processes the video generating captions from small video clips. A LLM iteratively refines the search by focusing on the most promising clips using the captions. The exploration follows a hierarchical reasoning: initially, a broad set of low-resolution frames is analyzed by the VLM for each promising clip, progressively narrowing down to fewer frames with higher resolution as the search concentrates on smaller temporal clips. The LLM uses captions, question semantics, answer completion, and confidence scores to determine whether to continue exploring or end the exploration. Once a high-confidence answer is found, or the search reaches a predefined threshold, FALCONEye outputs the final answer, its confidence score, and the corresponding temporal interval.}
    \label{fig:teaser}
\end{figure*}

\begin{abstract}
Finding information in hour-long videos is a challenging task even for top-performing Vision Language Models (VLMs), as encoding visual content quickly exceeds available context windows.
 To tackle this challenge, we present \textbf{FALCONEye}\footnote{\url{https://cplou99.github.io/FALCONEye/}}, a novel video agent based on a training-free, model-agnostic meta-architecture composed of a VLM and a Large Language Model (LLM). FALCONEye answers open-ended questions using an exploration-based search algorithm guided by calibrated confidence from the VLM’s answers. We also introduce the FALCON-Bench benchmark, extending  Question Answering problem to \textbf{V}ideo \textbf{A}nswer \textbf{S}earch—requiring models to return both the answer and its supporting temporal window for open-ended questions in hour‑long videos. 
With just a 7B VLM and a lightweight LLM, FALCONEye outscores all open‑source 7B VLMs and comparable agents in FALCON-Bench. It further demonstrates its generalization capability in MLVU benchmark with shorter videos and different tasks, surpassing GPT‑4o on single‑detail tasks while slashing inference cost by roughly an order of magnitude.
\end{abstract}
    
\section{Introduction}
\label{sec:intro}

Vision Language Models (VLMs) have demonstrated remarkable performance on complex, fine-grained questions over images and short videos~\cite{llava-onevision, Qwen2.5-VL, openai2023gpt4v}. However, extending these capabilities to long-form videos remains an open challenge~\cite{chandrasegaran2024hourvideo, zhou2024mlvu, xiao2025videoqa}. Our work emphasis is on analyzing very long videos and single-detail questions whose answers are contained on a tiny set of frames, \textit{a needle in a haystack} problem.  Not only that, we go a step further by pushing the task boundaries—focusing on open-ended questions and requiring models to return both the answer and visual evidence of the answer in the form of a temporal window  where the answer is found. This forces the agent to perform a deeper reasoning, limiting the number of hallucinations or random guesses deduced purely from the question or options without visual confirmation~\cite{rana2025mvu}. We refer to this extended setting as \textbf{Video Answer Search (VAS)} task.

Unlike general video understanding—where a global, embedded representation of the entire video may suffice for coarse-level tasks—VAS demands precise information retrieval within a large temporal space. Current VLMs typically address long video inputs by uniformly sampling frames (typically in the order of dozens or hundreds) until reaching the context window size limit \cite{zhang2024llavanext-video, ye2024mplug}. While this strategy may support high-level summarization or scene classification, it often fails when the answer lies in a small, temporally-localized segment~\cite{xiao2025videoqa}. In VAS, locating a single relevant moment among thousands of frames requires not just compression, but active reasoning and targeted exploration—capabilities that current VLMs still lack.

% EXISTING WORK
Recent efforts have focused on fine-tuning VLMs for long-form video understanding~\cite{park2024too, wang2025videotree, shen2024longvu, song2024moviechat, longvila, Qwen2.5-VL, shu2025video, shu2025video}. These models aim to encode more frames by optimizing sampling and token compression strategies to fit within the LLM's context window. While this enables a more complete global representation of the video, it comes at the cost of temporal resolution and precision—making them insufficient for tasks like VAS, where identifying a short, specific segment is crucial. To bridge this gap, several LLM-based agents such as such as VideoAgent \cite{fan2024videoagent} or VideoTree~\cite{wang2025videotree} use meta-architectures combining multiple models to extract key frames—often via CLIP features—and answer multiple-choice questions.

In contrast, we introduce \textbf{FALCONEye}, a novel video agent based on a training-free, model-agnostic meta-architecture composed solely of a VLM and an LLM. It is capable of answering single-detail, open-ended questions in hour-long videos by following a video search algorithm guided by calibrated confidence from the VLM’s outputs—avoiding irrelevant clips and focus computational resources on exploring the most promising clips.
Figure \ref{fig:teaser} illustrates FALCONEye's dynamic and iterative approach.

Overall, our work presents the following contributions:
1) \textbf{FALCON-Bench}: The first benchmark specifically designed for  VAS task. It focuses on open-ended (OE) questions that require locating answers in tiny temporal regions of one-hour-long videos, filling the challenging gap left by existing VQA benchmarks regarding accuracy and temporal answer localization for OE questions in longer videos.\\
2) \textbf{Novel video agent (FALCONEye)}: A system to emulate human-like behavior to answer single-detail questions in videos,  integrating VLM and LLM capabilities in a training-free and model agnostic meta-architecture. It leverages our novel search algorithm, that reasons to iteratively focus on video clips most likely to contain the answer. FALCONEye and FALCON-Bench have been built under the \texttt{lmms-eval} framework \cite{zhang2024lmmseval} to facilitate reproducibility and future evaluations. They are a significant advancement in VAS and, VQA in general, setting a new standard for precision-guided exploration in long-form video.  \\
3) \textbf{Evaluation of state-of-the-art VLMs calibration}: A study assessing the calibration of leading VLMs in VQA.
\section{Related work}

\paragraph{Vision Language Models (VLMs).}
VLMs emerged to extend LLMs' capabilities to vision, with the seminal work of LLaVA~\cite{liu2023llava}, which integrated a pre-trained CLIP visual encoder (ViT-L/14~\cite{radford2021learning}) to extract visual tokens, with an LLM (Vicuna~\cite{chiang2023vicuna}) for multimodal text and image understanding. This integration appears in other open-source architectures which improved the alignment and processing of visual features, such as BLIP-2~\cite{li2023blip}, LLaVA-1.5~\cite{liu2023improvedllava}, mPLUG-Owl2~\cite{ye2024mplug}, and LLaMa3.2-Multimodal~\cite{llama3_2}. 

More recently, Qwen2-VL~\cite{Qwen2-VL} introduced a VLM with a ViT capable of processing images at arbitrary resolutions, dynamically converting them into a variable number of visual tokens. 
Many companies have integrated increasingly powerful VLMs into their proprietary long-context chatbot systems, including  OpenAI’s GPT-4V~\cite{openai2023gpt4v}, Google DeepMind’s Gemini~\cite{google2023gemini} or Anthropic’s Claude~\cite{anthropic2024claude}. VLMs excel at image questions, but extending these capabilities to videos remains an open challenge due to the large number of tokens required to represent video content.

\textbf{VLM strategies for video}
extend VLMs to process multiple frames as input~\cite{lin2023video, llava-next, li2024mvbench, llava-onevision, zhang2024llavanext-video}  employing a uniform video sampling strategy to extract and process up to 64 frames, to keep the  visual token amount within the LLM’s context window. Specific \textbf{long-video} solutions target efficient and sophisticated video representations. 
MovieChat~\cite{song2024moviechat} builds a global video representation combining short- and long-term token-based memories,  \textcolor{black}{Video-XL-Pro} \cite{liu2025video} and LongVU~\cite{shen2024longvu} 
present a spatiotemporal adaptive compression to
retain the most relevant parts for question answering, and LongVILA~\cite{longvila} introduces Multi-Modal Sequence Parallelism
to extend the number of video frames processed from 8 to 2048. Apollo~\cite{zohar2024apollo} conducts an extensive study on strategies to enhance long-video understanding. \textcolor{black}{Differently, ReKV \cite{di2025rekv} is a training-free mechanism that retrieves relevant information from a video key-value cache to enable efficient streaming VQA.} 
Recently, Qwen2.5-VL~\cite{Qwen2.5-VL} extends its dynamic resolution capabilities to the temporal dimension, allowing processing up to 768 frames. 
Unlike these methods focused on answering by means of global complete video understanding, we focus on finding a short clip that contains the answer and returning the answer.

\begin{table}[!t]
\centering
\caption{\textbf{FALCON-Bench} compared to  popular video VQA benchmarks: number of videos (\#V) and questions (\#QAs); average video duration in seconds (s); multiple-choice (MC) or open-question (O) in the evaluation (underlined if one of them is dominant in the benchmark); inclusion of questions focused on single detail (SD) video aspects; provided GT time interval ($\Delta t$) where the answer is contained; VAS task evaluation. \textcolor{black}{A key distinction is wether QA and ground truth (QA-GT) are \textit{human} annotations or automatically extracted from available video description (Auto).}}

\resizebox{\columnwidth}{!}{%
\begin{tabular}{@{}l@{\hspace{1mm}}c@{\hspace{1mm}}c@{\hspace{1mm}}c@{\hspace{1mm}}c@{\hspace{1mm}}c@{\hspace{1mm}}c@{\hspace{1mm}}c@{\hspace{1mm}}c}
\toprule
\textbf{Benchmark} & QA-GT & \#V & \#QAs & \textbf{Dur.(s)} &  Eval. & SD & $\Delta t$ & \textbf{VAS} \\
\cmidrule(r){1-1} \cmidrule(r){2-9}
ActivityNet-QA~\cite{yu2019activitynet} & Human & 5,800 & 58,000 & 180 &   O & \xmark & \xmark & \xmark\\
MVBench~\cite{li2024mvbench} & Auto & 4,000 & 4,000 & 16 &  MC & \xmark & \xmark & \xmark\\
EgoSchema~\cite{mangalam2023egoschema} & Auto & 5,031 & 5,031 & 180 & MC & \xmark & \xmark & \xmark\\
Event-Bench~\cite{du2024towards} & Both &  1,294 & 2,190 & 101 &  MC & \cmark (21\%) & \xmark  & \xmark\\
LongVideoBench~\cite{wu2024longvideobench} & Human  & 3,763 & 6,678  & 473 &  MC & \cmark (30\%) & \xmark  & \xmark \\
{\textcolor{black} LongVideoBench-L~\cite{wu2024longvideobench} } & Human & 966 & -  & 1408 &  MC & \cmark (30\%) & \xmark  & \xmark  \\
Video-MME~\cite{fu2024video} & Human &  900 & 2,700 & 1,017 & MC & \cmark (15\%) & \xmark  & \xmark\\
{\textcolor{black} VIDEO MME-Long~\cite{fu2024video} } & Human &  300 & 900 & 2466,7 & MC & \cmark (15\%) & \xmark  & \xmark\\
HourVideo~\cite{chandrasegaran2024hourvideo} & Human & 500 & 12,976 & 2,742  & MC & \xmark & \xmark & \xmark \\
LVBench~\cite{wang2024lvbench}  & Human & 103 & 1,600 & 4,101 &  MC & \cmark (14\%) & \xmark  & \xmark\\
NeXT-QA~\cite{xiao2021next} & Human & 5,440 & 52,044 & 44 &  MC/O & \xmark & \xmark & \xmark\\
MovieChat-1K~\cite{song2024moviechat} & Human & 130 & 1,950 & 500 & MC/\textbf{\underline{O}} & \xmark & \xmark & \xmark\\
{\textcolor{black} MovieChat-1K $>10'$~\cite{song2024moviechat}} & Human & 19 & - & $>600$ & MC/\textbf{\underline{O}} & \xmark & \xmark & \xmark\\
CG-Bench~\cite{chen2025cgbench} & Human &  1,219 & 12,129 & 1,624 &  \underline{MC}/O & \cmark (100\%) & \cmark  & \xmark\\ 
MLVU~\cite{zhou2024mlvu} & Human &  1,730 & 3,102 & 930 &  MC/O & \cmark(64\%) & \xmark  & \xmark\\
{\textcolor{black} MLVU $>30'$~\cite{zhou2024mlvu} } & Human &  125 & - & $>1800$ &  MC/O & \cmark(64\%) & \xmark  & \xmark\\
InfiniBench~\cite{ataallah2024infinibench} & Auto & 1,219 & 108,200 & 3,155 &  MC/O & \cmark (14\%) & \xmark  & \xmark\\
{\textcolor{black} QAEgo4D~\cite{Baermann_2022_qaego4d} } & Auto &  1,325 & 14,513 & 495,1 & MC/O & \cmark (100\%) & \cmark & \cmark \\
\midrule
\textbf{FALCON-Bench} & \colorbox{lightgray}{Human} & 92 & 576 & \colorbox{lightgray}{\textbf{4,738}} & MC/\colorbox{lightgray}{\textbf{\underline{O}}} & \colorbox{lightgray}{\cmark (100\%)} & \colorbox{lightgray}{\cmark} & \colorbox{lightgray}{\cmark}\\
\bottomrule
\end{tabular}
}
\label{tab:benchmarks_related}
\end{table} 

\textbf{VQA Benchmarks.}
Existing Video QA benchmarks evaluate VLMs capabilities for plenty of video comprehension tasks. 
Early benchmarks focus on short videos and questions
 to understand temporal actions~\cite{xiao2021next},  human actions \cite{yu2019activitynet}, egocentric videos~\cite{mangalam2023egoschema} or movies~\cite{li2024mvbench}.
Similar to ours, Event-Bench~\cite{du2024towards} focuses on single-detail events occurred within short temporal windows but uses one-minute-long videos. Recently, several VQA benchmarks~\cite{chandrasegaran2024hourvideo, wu2024longvideobench, zhou2024mlvu, fu2024video}
extend the temporal duration of videos to around 10 minutes. Unlike these benchmarks, primarily focused  on Multiple-Choice Questions (MCQs), MovieChat-1K~\cite{song2024moviechat} is centered on open-ended questions (OQs). LVBench~\cite{wang2024lvbench} and InfiniBench~\cite{ataallah2024infinibench} further push this challenge with longer videos, up to 1 hour. Some of these later benchmarks include a percentage of single-detailed questions, with the answer located  within a short video time window, %like Video-MME, Event-Bench, MLVU and LVBench, 
but only LongVideoBench~\cite{wu2024longvideobench} and InfiniBench evaluate a small fraction of the questions in open-ended mode. Recently, CG-Bench~\cite{chen2025cgbench} introduced a benchmark focused on single-detail questions and medium-duration videos (average of 26 minutes per video), being the first to provide ground-truth time intervals for answers and to  evaluate state-of-the-art VLMs on the accuracy and temporal localization separately. {\textcolor{black} QAEgo4D~\cite{Baermann_2022_qaego4d} also introduces a task close to VAS in egocentric videos based on the Ego4D NQL narrations and annotations~\cite{grauman2022ego4d}. However, it contains  videos of average length 8 minutes and  questions automatically generated from the narration, usually focused on the main actor or object being interacted.}
In contrast, our \textbf{FALCON-Bench} takes the challenge a step further: it moves to one-hour-long videos, focuses on open-ended questions, and evaluates  the VAS task—requiring models to jointly return both the answer and its corresponding supporting temporal window.
Table~\ref{tab:benchmarks_related}  highlights {\textcolor{black}  FALCON-Bench more challenging (longer and OE focused) benchmark for VAS task, including Human annotations.}

\textbf{VQA Agent–style meta-architectures.}
Recent works have built agents for video QA combining an LLM with other architectures (e.g., VLMs, feature extractors). 
On one hand, there are \textit{fine-tuned solutions}, e.g., \emph{TimeChat}~\cite{ren2024timechat} fine-tunes a VLM with timestamp-aware encoder and sliding video Q-Former; or \emph{SeViLA}~\cite{yu2023self} chains a fine-tuned BLIP-2 to localise frames before answering.  
On the other hand, there are very successful \textit{training-free pipelines.}  
\emph{MoReVQA}~\cite{min2024morevqa} and \emph{ProViQ}~\cite{choudhury2024video} assemble multi-stage programs with ViperGPT~\cite{suris2023vipergpt} modules, while \emph{MVU}~\cite{ranasinghe2024understanding} \textcolor{black}{and VideoAgent2 \cite{ranasinghe2024understanding}} inject object-centric tracks into the LLM.  
\emph{Long Story Short}~\cite{chung2023long} uses a VLM to caption clips, a LLM to select the most relevant ones for a given MCQ and visual-text cosine similarity from CLIP \cite{radford2021clip} features to choose. the predicted option.  
\textcolor{black}{Similarly, LifelongMemory \cite{wang2024lifelongmemory} leverages an LLM to filter pre-extracted captions and answer MCQs from them.}
\emph{VideoAgent}~\cite{wang2024videoagent} loops over (i) CLIP-based frame selection (ii) VLM captions of key frames, (iii) LLM answers a MCQ from captions with an \emph{uncalibrated} confidence score until reaching a confident answer. VideoAgent\cite{fan2024videoagent} couples a VLM captioner \cite{zhao2023learning}, a ViCLIP video encoder \cite{wang2023internvid}, a GPT‑based LLM, an object detector\cite{zhao2024detrs} and an object tracker \cite{zhang2022bytetrack} to tackle VQA.
\emph{VideoTree}~\cite{wang2025videotree} builds a hierarchical tree of key frames (pre-extracted with EVA
CLIP-8B \cite{sun2024eva}), captions each node with a VLM and queries the LLM for the final answer.
The only agent that only requires a VLM and an LLM is the \emph{Socratic} baseline~\cite{zeng2022socratic, zhang2023simple}, where an LLM reasons only over VLM-generated captions. In this line, FALCONEye
(i) is a training-free and model-agnostic meta-architecture that only requires a VLM and an LLM;
{\textcolor{black}(ii) is the \emph{first} video agent to bring focus on strategies particularly designed for the more challenging \emph{open-ended} VQA;}
(iii) its exploration algorithm is guided by a calibrated confidence estimate derived from the VLM’s open answer, not by an ad-hoc integer from the LLM;  
(iv) is powered by an adaptive search  at clip-level: the agent varies clip length, frame count and resolution on-the-fly instead of relying on key-frame extraction \textcolor{black}{; (v) unlike existing agents, relying on dense captioning containing the answer, it  reasons from very light captions to generate the answer directly from the target visual content.}

\section{FALCONEye}\label{Sec:method}
FALCONEye is a novel video agent combining a training-free and model-agnostic meta-architecture (VLM+LLM) with an exploration-based search algorithm guided by calibrated confidence from VLM's answers. Unlike previous agents designed for MCQs, short videos and frame-level operations, FALCONEye handles OQs, one-hour-long videos, and operates at clip level.

\begin{figure*}[!t]
    \centering
    \includegraphics[width=0.8\linewidth]{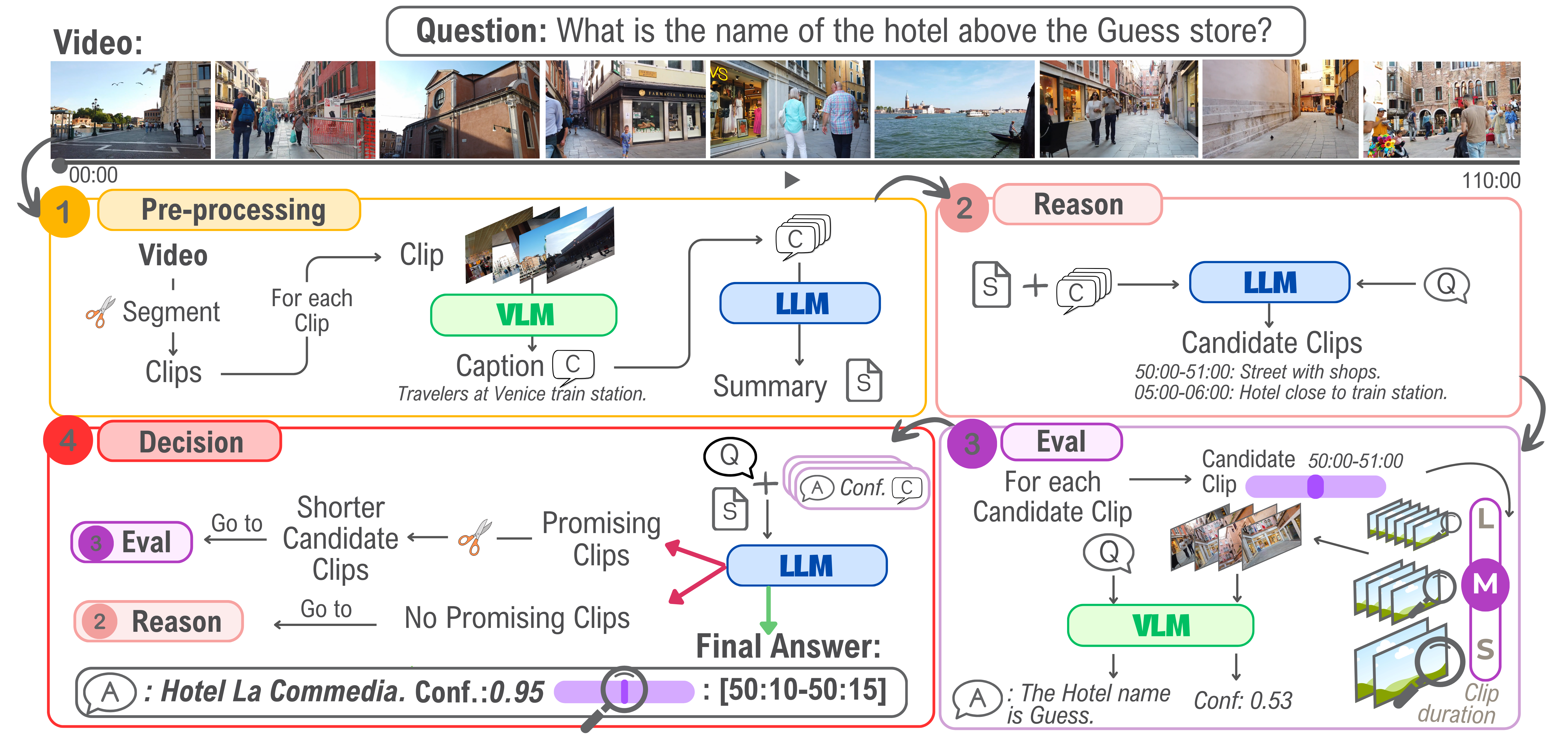}
    \caption{\textcolor{black}{\textbf{FALCONEye exploration algorithm.}}
    %that follows our FALCONEye meta-architecture.} 
    Given a question (\hspace{-0.5mm} \raisebox{-0.25em}{\includegraphics[height=1em]{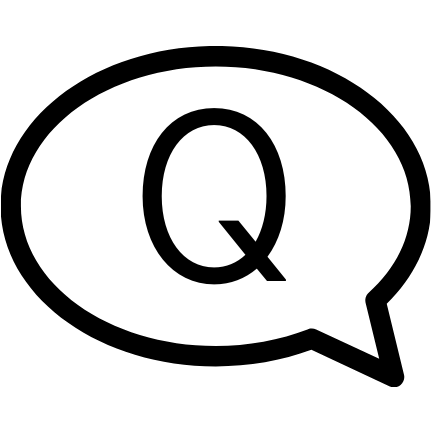}}) and a video, it starts with a global \textbf{Pre-processing} of the video, where the VLM generates short captions (\hspace{-0.5mm} \raisebox{-0.25em}{\includegraphics[height=1em]{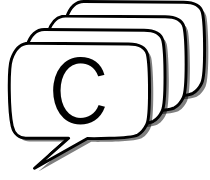}}) from video clips. Afterwards, the LLM \textbf{Reasons} from the Q and the captions to select some candidate clips to explore. The VLM \textbf{Evaluates} the candidate clips sampling frames varying number and resolution according to the clip duration. We get an answer (\hspace{-0.5mm} \raisebox{-0.25em}{\includegraphics[height=1em]{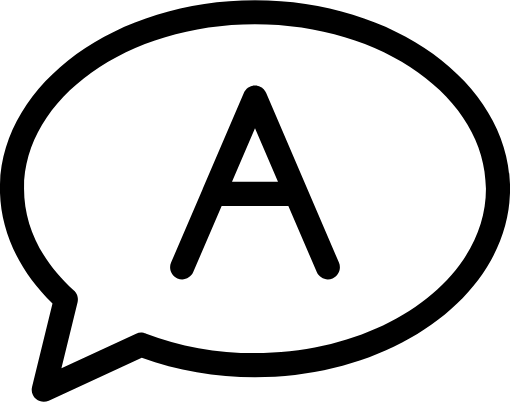}}) and its confidence (\textcolor{black}{\emph{Conf.}}) for each candidate clip. From all \{\hspace{-0.5mm} \raisebox{-0.25em}{\includegraphics[height=1em]{images/Answer.png}}, \textcolor{black}{\emph{Conf.}}, \hspace{-1.5mm} \raisebox{-0.25em}{\includegraphics[height=1em]{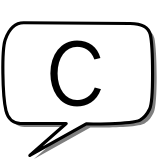}}\} tuples, the LLM \textbf{Decides} wether the final answer has already been found or \textcolor{black}{to continue exploring with two different ways: if some candidate clips are still promising and large enough, we segment them to generate new shorter candidate clips to evaluate. Otherwise, we ask again the LLM to reason removing the caption of the already explored clip.}}
    \label{fig:FALCONEyeExploration}
\end{figure*}
\vspace{-0.4cm}
\paragraph{VAS Problem definition:}
Given a video-question pair ($\mathcal{V}$, $Q$), the goal of the VAS task is:
\begin{enumerate}
    \item Find the answer $\mathcal{A}$ that accurately responds to $Q$.
    \item Localize a short video clip $v \in \mathcal{V}$ as evidence, ensuring that the answer $\mathcal{A}$ can be observed on its frames.
\end{enumerate}

\subsection{Meta-architecture}
We propose a meta-architecture with two main components: a VLM and an LLM. This meta-architecture is training-free and agnostic to any specific LLM or VLM, allowing flexibility in model selection.
Our overall idea involves using a small VLM to process the most computationally expensive data (visual information) and a medium-sized LLM to handle text-based reasoning, which is significantly cheaper to process. This setup enables higher efficiency and cost reduction compared to addressing the task with current large-scale state-of-the-art VLMs.
The VLM is used only to process short video clips, namely for two tasks, provide captions and generate potential answers to $Q$. The LLM powerful reasoning capabilities are exploited to extract information from all the captions and find interesting clips, and to decide wether to explore other clips, refine the current clip search (zooming into it) or stop the search.

\subsection{Exploration Algorithm}\label{Sec:explorationalgorithm}
The exploration algorithm consists of four main stages: \textbf{\circled{1} Pre-processing}, \textbf{\circled{2} Reasoning}, \textbf{\circled{3} Evaluation}, and \textbf{\circled{4} Decision} (Figure~\ref{fig:FALCONEyeExploration}). First, a pre-processing phase, which can be done offline for any video as it is question independent. For question-answering, stages \circled{2}-\circled{4} are executed in a loop, which ends when either the LLM selects a potential answer as the final one, the LLM reasons that there are no more candidate clips to explore, or a maximum number of iterations is reached. Algorithm~\ref{alg:FALCONEye} and Section~\ref{Sec:Prompts} in the supplementary material detail the pseudo-code of this algorithm and all the prompts used to interact with the LLM.

\textbf{$\circled{1}$ Pre-processing-.}
Here, our method generates short captions and a concise summary of the video. The video $\mathcal{V}$ is segmented into short video clips $\{v_1,\ldots,v_n\}$, that we call \emph{first-level clips}, such that $\mathcal{V}=\cup_{i=1}^{n}v_{i}$. For each video clip $v_i$, the VLM uniformly samples a small set of frames to generate a short caption $c_i$ that summarizes its content. Given the set of all captions $\mathcal{C}=\{c_1,\ldots,c_n\}$, the LLM then generates a global yet brief summary $\mathcal{S}$ of the video.

\textbf{\circled{2} Reasoning-.} 
Given the summary $\mathcal{S}$, the set of captions $\mathcal{C}$, and the question $Q$, the LLM pinpoints a set of first-level clips $\{v^{*}_1,\ldots,v^{*}_m\} \subset \mathcal{V}$ as \emph{candidates} to contain the answer. The number of selected clips $0 \leq m \leq n$ is directly selected by the LLM and it varies each iteration. If the LLM determines that no clip is likely to contain the answer, i.e.: $m=0$, the process terminates.

\textbf{\circled{3} Evaluation-.} 
In this stage, each candidate clip $v^{*}_i$ is evaluated by the VLM. First, a uniform sampling of frames is performed within the clip. The number and resolution of sampled frames vary according to the duration of $v^{*}_i$, creating a zoom-in effect. For first-level clips, a larger number of frames with lower resolution are sampled, whereas for shorter clips, fewer frames with higher resolution are selected. For fixed input resolution VLMs, we use tiling to increase the resolution (similar to what GPT4o~\cite{openai2023gpt4v} does). 

The VLM takes as input the question $Q$ along with the sampled frames from $v^{*}_i$ and generates a \textbf{potential answer} $\mathcal{A}^{*}_i$ and a \textbf{confidence score} $p(\mathcal{A}^{*}_i)$. The answer is derived from a set of generated tokens $\mathcal{T} \in D$, such that $\mathcal{A}^{*}_i=Tokenizer(\mathcal{T})$, where $D=\{t_1,\ldots, t_d\}$ is the token dictionary used by the LLM. 
During the VLM generation process, we store the logits $\{l_{1}^k,\ldots, l_{d}^k\}$ for each generated token $t_k \in \mathcal{T}$ and compute $p(\mathcal{A}^{*}_i)$ as the geometric average of the answer token probabilities~\cite{liu2023litcab}:%
\begin{equation}\label{eq:geom_avg}
\footnotesize
    p(\mathcal{A}^{*}_i) = \sqrt[s]{\prod_{k=1}^{s}\max \left( Softmax\left(\{l_{1}^k,\ldots, l_{d}^k\}\right) \right)},%
\end{equation}%
where $s$ is the number of generated tokens in $\mathcal{T}$  (see Fig.~\ref{fig:probsFALCON} in supp. material). In MCQ scenario, there is no need for an average as $s=1$ (a single token for the option letter: A, B, C, D, \ldots) and the confidence is just the \emph{softmax} of the logits for the option letter token~\cite{chung2023long}. This step is supported by our VLM calibration analysis presented later in Section~\ref{Sec:CalibrationStudy}.

\textbf{\circled{4} Decision-.}
At this stage, the LLM receives the information of the candidate clips that have just been evaluated in the last iteration of the Evaluation stage, $\{\mathcal{A}^{*}_i, p(\mathcal{A}^{*}_i), c_i \}$, along with the summary $\mathcal{S}$ and the question $Q$. Based on confidence values, answer quality, captions, and the temporal localization of the clips, the LLM determines whether one of these answers should be selected as the final answer. If an answer meets the selection criteria, the process \textbf{returns the final answer} $\mathcal{A}$, its temporal localization, and its confidence $p(\mathcal{A})$. Otherwise, the system identifies which of the candidate clips $\{v^{*}_1,\ldots,v^{*}_m\}$ remain \emph{promising} for further exploration and zoom-in. The number of promising clips is not fixed and varies based on the LLM's judgment.  If the identified \emph{promising clips} are first-order or second-order, they are segmented into shorter clips, converting them into second-order and third-order clips, respectively. These new clips form an updated set of candidate clips $\{v^{*}_1,\ldots,v^{*}_m\} \subset \mathcal{V}$, and the process \textbf{returns to Stage \circled{3} Evaluation}.  If the selected \emph{promising clips} are at the third-order level, the system \textbf{returns to Stage \circled{2} Reasoning}. To prevent redundant exploration, captions corresponding to previously evaluated first-order clips are removed from the caption set $\mathcal{C}$.

If the loop between stages \circled{2} and \circled{4} ends without producing a final answer, a final decision must be made. This situation occurs when either the maximum number of iterations—defined as the number of candidate clips evaluated by the VLM—has been reached, or when the LLM determines that no more first-order clips remain as candidates for exploration.  In this case, the LLM gathers the information from all previously evaluated clips $\{\mathcal{A}^{*}_i, p(\mathcal{A}^{*}_i), c_i \}$ and selects the most suitable answer among them. Finally, the process outputs the final answer $\mathcal{A}$, its temporal localization, and its confidence value $p(\mathcal{A})$.

\section{FALCON-Bench}

\textbf{FALCON-Bench} is the first benchmark specifically tailored to the VAS task, where models must not only provide an answer to a question, but also retrieve the precise temporal window that supports it.  Our benchmark raises the difficulty across all axes: we use one-hour-long videos, we focus on open-ended questions, and evaluate both answer quality and temporal grounding jointly. 
While all videos were obtained from external public datasets, our contribution lies in the constructed set of Q\&A that assess the VAS task on these videos, where each question is carefully crafted to ensure a unique answer within the video, with the answer content embedded in a single, short temporal window. 
Unlike previous VQA benchmarks, which often rely on broad video-level reasoning, we highlight the challenge of finding a \textit{needle in a haystack} within long-form videos. Our benchmark emphasizes the necessity of not just uniformly sampling frames or applying token compression across the entire video, but instead reasoning over the content to localize the exact clip that contains the answer.

The videos of our benchmark were sourced from \textbf{S}occerNet \cite{giancola2018soccernet}, \textbf{M}ovieChat-1K \cite{song2024moviechat} and Walking \textbf{T}ours Dataset \cite{venkataramanan2023imagenet}, ensuring diversity and comprehensive coverage across different video types. Overall, the benchmark comprises 575 questions, covering 4 categories, over 90 videos, with an \textbf{average video duration of 78.9 minutes} and \textbf{answers localized within a GT temporal window of 38.4 seconds}. The dataset is split into a test set (506 questions) and a validation set (70 questions). 
Each question is accompanied by four different answer options, allowing for multiple-choice question (MCQs) evaluation. However, the primary focus remains on open-ended questions (OQs), as this approach is more aligned with the VAS task. The annotation process was conducted by humans assisted by GPT-4o for generating answer options. More details on the benchmark content and construction process are in Sec.~\ref{Sec:bench-details} of the supplementary material. The benchmark has been built under the \texttt{lmms-eval} framework \cite{zhang2024lmmseval} to ease further evaluation and will be publicly available upon acceptance.

\begin{table*}[!t]
\centering
\caption{Models performance comparison in FALCON-Bench test split, both MCQs and OQs. We provide LLM size (S), mean number of frames (\#F) processed per question by each VLM. Results grouped by MovieChat (M), Soccer (S), WalkingTours (T), and average (Avg.). 
}
\label{tab:FALCON-BenchMainResults}
\resizebox{\textwidth}{!}{%
\begin{tabular}{l cc cccccc cccccccccc}
\toprule
\multicolumn{3}{c}{Model} & \multicolumn{6}{c}{\textbf{Multiple-Choice Questions}} & \multicolumn{10}{c}{\textbf{Open Questions}}  \\
 \cmidrule(r){1-3} \cmidrule(r){4-9} \cmidrule(r){10-19}
 Name &  \multicolumn{2}{c}{Config.} & Time &  \multicolumn{4}{c}{Accuracy} & Loc. & Time & \multicolumn{4}{c}{Accuracy} & \multicolumn{4}{c}{Score} & Loc. \\
 \cmidrule(r){1-1} \cmidrule(r){2-3} \cmidrule(r){4-4} \cmidrule(r){5-8} \cmidrule(r){9-9} \cmidrule(r){10-10} \cmidrule(r){11-14} \cmidrule(r){15-18} \cmidrule(r){19-19}
  & S &  \#F & s  & M & S & T & Avg. & mGToU & s & M & S & T & Avg. & M & S & T & Avg. & mGToU \\
\midrule

\hline
\rowcolor{gray!10}  \multicolumn{19}{|l}{\emph{\textbf{Baselines}}} \\
\hline
Full Mark & - & - & - & 100 & 100 & 100 & 100 & 100 & - & 100 & 100 & 100 & 100 & 5.00 & 5.00 & 5.00 & 5.00 & 100 \\
Human & - & - & 115.6 & 100 & 100 & 76.9 & 92.3 & 73.3 & 129.2 & 84.6 & 95.0 & 76.5 & 85.4 & 4.31 & 4.80 & 4.06 & 4.42 & 69.2\\
GPT-4o-mini-blind & N/A & - & 0.48 & 45.0 & 32.4 & 42.0 & 39.8 & 0.00 & 1.76 & 8.82 & 0.00 & 2.00 & 3.60 & 0.48 & 0.00 & 0.18 & 0.22 & 0.00 \\
Qwen2.5-VL-blind & 7B & - & 0.08 & 42.0 & 34.1 & 48.0 & 41.4 & 0.00 & 2.45 & 1.96 & 0.28 & 2.00 & 1.41 & 0.21 & 0.11 & 0.30 & 0.20 & 0.00 \\
Random & - & - & 0.00 & 25.0 & 25.0 & 25.0 & 25.0 & 0.00 & 0.00 & 0.00 & 0.00 & 0.00 & 0.00 & 0.00 & 0.00 & 0.00 & 0.00 & 0.00 \\
\hline
\rowcolor{gray!10} 
\multicolumn{19}{|l|}{\emph{\textbf{Open-Source VLMs with Multi-Image Support}}} \\
\hline
mPLUG-Owl3 \cite{ye2024mplug} & 7B & 16 & 5.52 & 41.1 & 28.5 & 36.0 & 35.2 & 0.00 & 6.63 & 8.82 & 4.23 & 10.0 & 7.68 & 0.68 & 0.29 & 0.66 & 0.54 & 0.00 \\
LLaVA-v1.5  \cite{liu2023llava} & 7B & 32 & 11.4 & 23.5 & 31.9 & 44.0 & 33.1 & 0.00 & 12.7 & 3.92 & 4.80 & 2.00 & 3.57 & 0.36 & 0.31 & 0.24 & 0.30 & 0.00 \\
LLaVA-v1.6  \cite{llava-next} & 7B & 32 & 12.9 & 28.4 & 30.5 & 38.0 & 32.3 & 0.00 & 14.3 & 4.90 & 4.51 & 2.00 & 3.80 & 0.31 & 0.30 & 0.20 & 0.27 & 0.00\\
LLaVA-OV \cite{llava-onevision} & 7B & 32 & 16.9 & 48.0 & 30.2 & 46.0 & 41.4 & 0.00 & 7.29 & 9.80 & 4.23 & 16.0 & 10.0 & 0.53 & 0.25 & 1.00 & 0.59 & 0.00\\
Qwen2.5-VL \cite{Qwen2.5-VL} & 7B & 768 & 68.9 & 50.0 & 24.4 & 48.0 & 40.8 & 0.00 & 70.1 & 11.7 & 8.19 & 10.0 & 9.96 & 0.72 & 0.53 & 0.62 & 0.62 & 0.00  \\
\hline
\rowcolor{gray!10} 
\multicolumn{19}{|l|}{\emph{\textbf{Open-Source VLMs designed for videos}}} \\
\hline
Video-LLaVA \cite{lin2023video} & 7B & 8 & 3.37 & 28.4 & 26.8 & 34.0 & 29.7 & 0.00 & 4.50 & 6.86 & 5.08 & 8.00 & 6.64 & 0.56 & 0.37 & 0.68 & 0.53 & 0.00\\
VideoChat2-HD \cite{li2024mvbench} & 7B & 8 & 6.74 & 29.4 & 24.2 & 26.0 & 26.5 & 0.00 & 8.06 & 8.82 & 2.82 & 8.00 & 6.54 & 0.71 & 0.20 & 0.52 & 0.47 & 0.00 \\
LLaVA-Video \cite{zhang2024llavanext-video} & 7B & 32 & 12.4 & 50.0 & 38.9 & 58.0 & 48.9 & 0.00 & 13.0 & 10.7 & 11.0 & 12.0 & 11.2 & 0.76 & 0.66 & 0.86 & 0.76 & 0.00 \\

\hline
\rowcolor{gray!10} 
\multicolumn{19}{|l|}{\emph{\textbf{Open-Source VLMs specific for long videos}}} \\
\hline

%LongVILA \cite{lin2024vila} & 7B & 2048 &  &  &  &  &  &  &  & & \\
MovieChat-OV \cite{song2024moviechat} & 7B & 1024 & 62.3 & 43.1 & 38.7 & 46.0 & 42.6 & 0.00 & 69.3 & 9.80 & 4.80 & 10.0 & 8.20 & 0.53 & 0.25 & 0.66 & 0.48 & 0.00 \\
%LongVU \cite{shen2024longvu} \textcolor{red}{OOM} & 1fps &  &  &  &  &  &  &  & & \\
Apollo \cite{zohar2024apollo} & 7B & 0.5fps & 36.2 & 51.9 & 35.8 & 64.0 & 50.5 & 0.00 & 36.5 & 16.6 & 10.7 & 18.0 & 15.1 & 0.96 & 0.68 & 1.06 & 0.90 & 0.00\\
\textcolor{black}{Video-XL-Pro}  \cite{liu2025video} & \textcolor{black}{3B} & \textcolor{black}{256} &  \textcolor{black}{48.8} &  \textcolor{black}{52.9} & \textcolor{black}{33.8} &  \textcolor{black}{50.0} & \textcolor{black}{45.6} & \textcolor{black}{0.00} & \textcolor{black}{7B} & \textcolor{black}{17.6} & \textcolor{black}{7.62} & \textcolor{black}{14.0} & \textcolor{black}{13.0} & \textcolor{black}{1.08}  & \textcolor{black}{0.45} & \textcolor{black}{0.94} & \textcolor{black}{0.63} & \textcolor{black}{0.00}  \\

\textcolor{black}{LLaVA-OV+ReKV} \cite{di2025streaming} & \textcolor{black}{7B} & \textcolor{black}{0.5fps} &  \textcolor{black}{92.1} &  \textcolor{black}{65.6} & \textcolor{black}{40.1} &  \textcolor{black}{56.0} & \textcolor{black}{53.9} & \textcolor{black}{0.00} & \textcolor{black}{96.6} & \textcolor{black}{34.3} & \textcolor{black}{13.8} & \textcolor{black}{24.0} & \textcolor{black}{24.0} & \textcolor{black}{1.78}  & \textcolor{black}{0.80} & \textcolor{black}{1.40} & \textcolor{black}{1.32} & \textcolor{black}{0.00}  \\

\hline
\rowcolor{gray!10} 
\multicolumn{19}{|l|}{\emph{\textbf{Meta-architectures built from a LLM (GPT4o-mini) and a VLM (Qwen2.5-VL)}}} \\
\hline

Socratic \cite{zeng2022socratic}  & 7B+N/A & 0.5fps & 186.6 & 53.9 & 48.0 & 54.0 & 51.9 & 21.3 & 186.7 & 25.4 & 29.0 & 20.0 & 24.8 & 1.49 & 1.71 & 1.16 & 1.45 & 19.6 \\

\textcolor{black}{LifelongMemory}\cite{wang2024lifelongmemory} &  \textcolor{black}{7B+N/A}  & \textcolor{black}{0.5fps} &  \textcolor{black}{207.3} &  \textcolor{black}{45.0} & \textcolor{black}{41.5} &  \textcolor{black}{38.0} & \textcolor{black}{41.8} & \textcolor{black}{0.00} & \textcolor{black}{208.2} & \textcolor{black}{6.86} & \textcolor{black}{22.3} & \textcolor{black}{8.00} & \textcolor{black}{12.3} & \textcolor{black}{0.53}  & \textcolor{black}{1.37} & \textcolor{black}{0.56} & \textcolor{black}{0.82} & \textcolor{black}{0.00} \\

VideoAgent\cite{wang2024videoagent}+OQ & 7B+N/A & 9.2 & 45.0 & 48.0 & 52.2 & 36.0 & 45.4 & 19.1 & 43.8 & 8.82 & 24.8 & 8.00 & 13.8 & 0.56 & 1.38 & 0.44 & 0.79 & 13.3 \\
\textbf{FALCONEye}\textcolor{black}{-Pro} & 7B+N/A & 659.2 & 393.0 & 74.5 & 61.5 & 74.0 & \textbf{70.0} & \textbf{27.7} & 245.4 & 50.9 & 37.2 & 46.0 & \textbf{44.7} & 2.67 & 1.96 & 2.50 & \textbf{2.38} & \textbf{24.9}  \\ 
\textbf{FALCONEye}\textcolor{black}{-Flash} & 7B+N/A & 518.5 & 213.3 & 67.6 & 58.4 & 68.0 & 64.7 & 25.2 & 187.9 & 43.1 & 38.4 & 42.0 & 41.1 & 2.36 & 2.03 & 2.28 & 2.22 & 22.7 \\ 

\bottomrule
\multicolumn{18}{l}{
{\small N/A means \textit{Not Available}, due to unknown model size or capability not available in the approach. -- means \textit{Does not apply}.}}\\
\end{tabular}
}
\end{table*}

\section{Experiments}
\paragraph{Metrics-.} In MCQs, we report \textit{accuracy} as the primary evaluation metric. For OQs, following~\cite{maaz2023video}, also applied in others, as InfiniBench~\cite{ataallah2024infinibench}, we employ GPT-assisted evaluation. GPT reports \textit{accuracy} (100 or 0) and a \textit{score value}  (0-5 range), assessing key aspects (Correctness of Information, Detailed Orientation, Contextual Understanding, Temporal Understanding and Consistency). 
We also use \textit{Ground Truth over Union} (GToU) metric to evaluate the quality of the predicted temporal  interval, as visual evidence (short clip) that contains the answer. Section~\ref{Sec:GToU} of the Supplementary discusses why we find GToU more adequate than mIoU (a clip that tries to match the entire GT clip).

\begin{table}[!b]
\centering
\caption{Meta-architectures cost and performance comparison in the test split of the FALCON-Bench OQs. We set LLM=GPT-4o-mini, VLM=Qwen2.5-VL. We measure LLM/VLM usage as number of inferences (\#Inf.) per question and, total tokens (\#Toks) and time (s) per inference. We report the average accuracy (Acc.), score (Sc.) and localization (Loc.).}
\label{Tab:FALCON-BenchMetaArch}

\resizebox{\columnwidth}{!}{%
\begin{tabular}{l ccc ccc c ccc}
\toprule
 & \multicolumn{3}{c}{LLM} & \multicolumn{3}{c}{VLM} & Total & \multicolumn{3}{c}{\textbf{Open Questions}}  \\
 \cmidrule(r){2-4} \cmidrule(r){5-7} \cmidrule(r){8-8} \cmidrule(r){9-11}

 Model Name & \#Inf. & \#Toks & s & \#Inf. & \#T & s  &  s & Acc. & Sc. & Loc. \\
\cmidrule(r){1-1} \cmidrule(r){2-4} \cmidrule(r){5-7}  \cmidrule(r){8-8} \cmidrule(r){9-11} 

Socratic  \cite{zeng2022socratic} &  1 & 13.5K & 2.1 & 12.6 &  768 & 14.6  & 186.7 & 24.8 & 1.45 & 19.6 \\
Socratic\textcolor{black}{-Short} \cite{zeng2022socratic} &  1 & 2.4K & 1.1 & 12.6 & 64 & 9.8 & 125.0 &  15.5 & 1.03 & 18.3 \\
\textcolor{black}{LifelongMemory}\cite{wang2024lifelongmemory} & \textcolor{black}{10.2} & \textcolor{black}{0.3K} & \textcolor{black}{1.0} & \textcolor{black}{12.6} & \textcolor{black}{64} & \textcolor{black}{9.8} & \textcolor{black}{208.2} & \textcolor{black}{12.3} & \textcolor{black}{0.82} & \textcolor{black}{0.00} \\
VideoAgent\cite{wang2024videoagent}+OQ & 5.6 & 1.3K & 3.2 & 3.70  & 64 & 2.2 & 43.8 & 13.8 & 0.79 & 13.3 \\
\midrule
SequentialBP & - & - & - & 53.3 & 64 & 12.7 & 676.9 &   28.7 & 1.52 & 13.7 \\
Sequential & - & - & - & 78.9 & 64 & 12.7 & 1002 &    34.8 & 1.83 & 9.05 \\
IterativeSamp & - & - & - & 85.4 & 64 & 18.7 & 1596 & 24.0 & 1.34 & 1.26  \\
\midrule
\textbf{FALCONEye}\textcolor{black}{-Pro} & 4.3 & 1.7K & 5.8 & 22.5 & 64 & 9.8 & 245.4 & 44.7 & 2.50 & 24.9 \\ 
\textbf{FALCONEye}\textcolor{black}{-Flash} & 3.0 & 1.7K & 5.8 &  17.4 & 64 & 9.8 & 187.9 & 41.8 & 2.24 & 22.7 \\ 
\bottomrule
\end{tabular}
}
\end{table}

\begin{figure}
     \centering
     \includegraphics[width=0.6\linewidth]{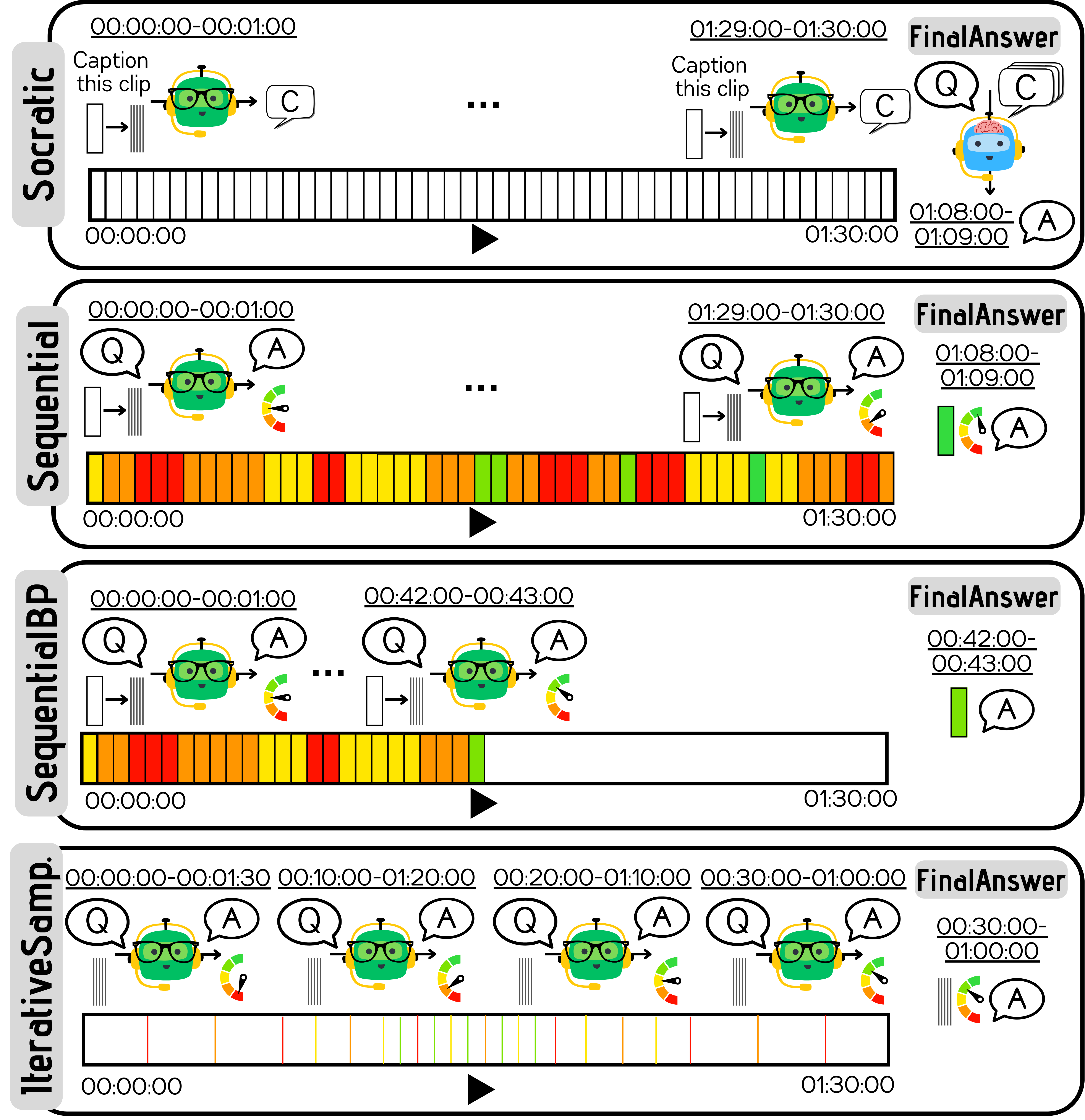}
     \caption{Visualization of the Socratic baseline approach together with our three exploration baselines considered to address VAS. }
     \label{fig:Baselines}
 \end{figure}

\vspace{-0.5cm}
\paragraph{Baselines-.}
A widely used baseline in recent long VQA benchmarks~\cite{chandrasegaran2024hourvideo} is the Socratic method~\cite{zeng2022socratic}. In this baseline, a VLM generates captions $\mathcal{C}=\{c_1,\ldots, c_n\}$ from short video clips $\{v_1,\ldots, v_n\} \in \mathcal{V}$ and a LLM generates $\mathcal{A}$ from $Q$ and $\mathcal{C}$. In addition to this baseline, we introduce three alternative exploration baselines (Fig.~\ref{fig:Baselines} of the supp. material) that rely solely on a VLM, guided by its answer confidence computed as described in Sec~\ref{Sec:method}. 

\begin{enumerate}
    \item \textbf{Sequential}: it sequentially processes the entire video $\mathcal{V}$ in short clips $\{v_1,\ldots, v_n\}$. For each short clip $v_i$, the VLM generates a potential answer $\{ \mathcal{A}^{*}_i, p(\mathcal{A}^{*}_i) \}$. At the end of the process, the final answer is selected as $A = \mathcal{A}^{*}_k$, where $k=\arg\max_i p(\mathcal{A}^{*}_i)$.
    
    \item \textbf{SequentialBP}: similar to the Sequential baseline, but with an early stopping criterion. The process halts as soon as an answer $\mathcal{A}^{*}_i$ is found with a confidence score $p(\mathcal{A}^{*}_i)$ exceeding a predefined threshold.
    
    \item \textbf{IterativeSampling}: it iteratively generates potential answers  within a progressively refined video clip $v^{*}$. The search begins with $v^{*}=\mathcal{V}$. The algorithm selects a small set of frames around the most unexplored regions within $v^{*}$ and the VLM generates $\{\mathcal{A}^{*}, p(\mathcal{A}^{*})\}$. Whenever a higher-confidence answer is found, $v^{*}$ is updated to the temporal window where the sampled frames were located, further narrowing the focus.
\end{enumerate}

As discussed in the related work, two relevant and recent video agents are VideoAgent~\cite{wang2024videoagent} and VideoTree~\cite{wang2025videotree}. 
We use  VideoAgent as our reference  meta‑architecture to compare with since it couples CLIP and a 7B‑parameter VLM, both of which fit comfortably on a single RTX3090. In contrast,  VideoTree  depends on EVA‑CLIP 8B—too large to co‑load with a VLM‑7B on that GPU—and its public codebase lacks the VLM captioning module, preventing a fair, fully reproducible baseline.

\vspace{-0.5cm}
\paragraph{FALCONEye-.}
We evaluate our meta-architecture using Qwen2.5-VL~\cite{Qwen2.5-VL} as the VLM and GPT4o-mini~\cite{openai2023chatgpt} as the LLM. As detailed in Sec.~\ref{Sec:CalibrationStudy}, Qwen2.5-VL outperforms other small-size VLMs on FALCON-Bench OQs in short clips near the answer, and is well calibrated. Its ViT supports images of any resolution, dynamically adjusting visual tokens, enabling an accurate zoom-in effect as clip lengths decrease. GPT4o-mini provides a balanced trade-off between cost and reasoning power (Sec.~\ref{Sec:FALCONEyeAblation}).  Prompts used for LLM queries are detailed in Sec.~\ref{Sec:Prompts}. Inspired by the Socratic approach~\cite{chandrasegaran2024hourvideo}, first-level clips are 1min length, but captions are now shorter formed by up to 64 tokens (instead of 768). Within these clips, second-order (20s) and third-order (5s) clips are explored. The zoom-in effect, validated in supplementary Sec.~\ref{Sec:zoom-in}, adjusts from 30 frames ($672\times1204$) in first-level clips to 10 frames ($824\times1462$) in third-order clips.  In OQs, the LLM determines the final answer based on completion, confidence ($0.8$ threshold), captions, and temporal localization (as explained in Sec.~\ref{Sec:method}). In MCQs, this evaluation is replaced by a simple confidence-based check with $0.9$ threshold since all possible answers are predefined (A, B, C, or D) and inherently valid, making confidence the only selection criterion. Lastly, we evaluate two versions of FALCONEye: FALCONEye\textcolor{black}{-Pro}, the top-performing configuration, which evaluates up to 45 candidate clips, and FALCONEye\textcolor{black}{-Flash}, a cost-efficient variant that limits evaluation to 10 candidates.

 \subsection{FALCON-Bench results}
Table~\ref{tab:FALCON-BenchMainResults} presents the performance of baselines, open-source VLMs, and meta-architectures on FALCON-Bench.  As baselines, we include human performance both to establish an upper bound and to analyze human strategies in VAS, which guided our algorithm design (Sec.~\ref{Sec:HumansExp}). Additionally, we evaluate our two selected LLMs (GPT4o-mini and Qwen2.5-VL) in blind mode (without visual information) to assess their reliance on text-based priors.  For open-source VLMs, we test small-scale versions of the main state-of-the-art models, including Apollo and MovieChat, which are specifically designed for long videos. We  also evaluate \textcolor{black}{three video agents}, Socratic~\cite{zeng2022socratic}, \textcolor{black}{LifelongMemory~\cite{wang2024lifelongmemory} for VQA} and VideoAgent~\cite{wang2024videoagent}. To ensure a fair comparison with our FALCONEye, these meta-architectures are implemented using the same VLM (Qwen2.5-VL) and LLM (GPT4o-mini). The Socratic approach generates captions (768 tokens) from 1-minute clips, while VideoAgent is extended to handle open-ended questions and predict answer localization, resulting in VideoAgent~\cite{wang2024videoagent}$+$OQ. Most VLMs that are limited to a small number of sampled frames fail to outperform the LLM-blind baselines in MCQs, suggesting that their advantage over random guessing stems from discarding certain options, rather than actual answer retrieval from visual content. A similar issue arises with Qwen2.5-VL when operating in its default mode (2fps, up to 768 frames), as it downscales frames to $140\times240$, losing crucial details. Only LLaVA-Video, Apollo, \textcolor{black}{and ReKV} show clear improvements over the blind baselines. This trend is not observed in OQs, where models must generate answers independently, validating their reliability.  Among meta-architectures, only FALCONEye demonstrates robustness for long-form VAS, \textbf{achieving 70.0\% accuracy in MCQs and 44.7\% in OQs} with its top-performance configuration, and 64.7\% and 41.1\%, respectively, in the cost-efficient variant. The latter configuration achieves performance close to the Socratic baseline while maintaining a processing time per question not so far from human response times. We analyze their costs and performances, including exploration baselines, in Table~\ref{Tab:FALCON-BenchMetaArch}. We also include Socratic\textcolor{black}{-Short}, which utilizes the short captions (64 tokens) generated by FALCONEye. While baseline exploration algorithms outperform existing meta-architectures, they require significantly more time. Notably, our LLM-guided exploration reduces this overhead while improving performance by efficiently narrowing the search.

\begin{table}[!t]
\centering
\caption{Model performance comparison on a VQA benchmarks with medium-duration videos — MLVU (720s), covering a diverse set of tasks beyond Single-Detail question types. FALCONEye surpasses comparable size/cost agents and GPT-4o at SD tasks.}
\label{tab:otherbenchmarks}
\resizebox{\columnwidth}{!}{%
\begin{tabular}{@{}l@{\hskip 3pt}*{15}{@{\hskip 3pt}c@{\hskip 3pt}}@{}}

\toprule
\textbf{MLVU} & \multicolumn{10}{c}{\textbf{Multiple-Choice}} 
& \multicolumn{3}{c}{\textbf{Generation}} \\
\cmidrule(r){2-11} \cmidrule(r){12-14} 

  & \multicolumn{4}{c}{\textbf{Single-Detail}} 
& \multicolumn{5}{c}{Non-Single-Detail} 
& \textbf{M-Avg} 
&  SD & NSD & \textbf{G-Avg} \\

\cmidrule(r){1-1} \cmidrule(r){2-5} \cmidrule(r){6-10} \cmidrule(r){11-11}  \cmidrule(r){12-12} \cmidrule(r){13-13} \cmidrule(r){14-14} 
\textbf{Model Name}  & NQA & ER & PQA & Avg. & TR & AR & AO & AC & Avg. & & SSC & VS & \\

\cmidrule(r){1-1} \cmidrule(r){2-5} \cmidrule(r){6-10} \cmidrule(r){11-11} \cmidrule(r){12-12} \cmidrule(r){13-13} \cmidrule(r){14-14}  

\hline
\rowcolor{gray!11}
\multicolumn{14}{|l|}{\emph{\textbf{Proprietary Long-Context LLMs}}} \\
\hline 
GPT-4o \cite{openai2023gpt4v} &  42.9 & 47.8 & 57.1 & 49.2 & 83.7 & 68.8 & 46.2 & 35.0 & \textbf{58.4} & \textbf{54.5} & 6.80 & 4.94 & \textbf{5.87} \\

\hline
\rowcolor{gray!11}
\multicolumn{14}{|l|}{\emph{\textbf{Meta-architectures built from a LLM (GPT4o-mini) and a VLM (Qwen2.5-VL-7B)}}} \\
\hline
VideoAgent \cite{wang2024videoagent} & 28.3 & 30.1 & 28.0 & 28.8 & 72.5 & 23.0 & 21.4 & 11.6 & 32.1 & 30.7 & 4.61 & 3.35 & 3.98 \\ 
\textbf{FALCONEye}\textcolor{black}{-Pro} & 51.6 & 45.2 & 52.0 & \textbf{49.6} & 80.2 & 30.7 & 27.1 & 3.33 & 35.3& 41.4 & 5.04 & 3.40 & 4.22 \\

\bottomrule
\end{tabular}}
\end{table}

\subsection{FALCONEye generalization evaluation}

To explore  \textbf{FALCONEye} generalization beyond our benchmark, we evaluate it \textit{out-of-the-box} on a standard VQA setting: shorter videos and a  broader mix of question types beyond single-detail (SD). With this goal, we select \textbf{MLVU}~\cite{zhou2024mlvu}, the most recent publicly available benchmark covering various VQA subtasks, not only  SD questions. A summary of these results is presented in Table~\ref{tab:otherbenchmarks}.

While MLVU does not reflect the two core challenges tackled by FALCONEye—namely, \emph{one-hour-long videos} (here, video durations are only a few minutes) and \emph{open-ended questions} (MLVU primarily includes multiple-choice questions)—it still offers valuable insights. Notably, shorter videos benefit standard VLMs like GPT-4o, as their frame sampling strategies (e.g., 256 frames) cover a large portion of the content, reducing the need for exploration. 
 
Despite this disadvantage, FALCONEye surpasses the SOTA video agent of comparable size across all subtasks, and even outperforms GPT-4o on SD questions, highlighting the strength of its exploration-driven, reasoning-based approach, even outside its native setting. 
Beyond performance, FALCONEye is also remarkably more cost‑efficient. A single question averages 7.3k tokens (1.7k tokens across 4.3 inferences) usage of GPT‑4o‑mini, costing roughly $\$0.01$ per question.
 Conversely, answering the same query with GPT‑4o and processing 256 frames ($\approx$ 22k tokens) costs about $\$0.11$ per question—nearly 10$\times$ more. This stark gap underscores FALCONEye’s scalability for long‑form video understanding.
\section{Configuration Analysis}\label{Sec:Ablation}

\paragraph{VLMs {\textcolor{black} confidence} calibration study.}\label{Sec:CalibrationStudy}
Since our exploration algorithm relies on answer confidence, here we evaluate if the confidence values are well calibrated, {\textcolor{black} i.e., the confidence correlates with the accuracy}. For that, we adopt Reliability Diagrams~\cite{degroot1983comparison}, which group predictions into bins based on confidence and measure the gap (calibration error) between confidence and accuracy for each bin. From these plots, we compute the Average Calibration Error (ACE) and Calibration Count (CC), quantifying the percentage of predictions above a defined confidence threshold, weighted by their accuracy calibration error (1-CE).  

Given the poor performance of state-of-the-art VLMs on FALCON-Bench, we ease the task by providing directly one-minute clips taken from the middle of the ground-truth (GT) temporal interval (Sec.~\ref{sec:falcon-bench-easy} of supp. material). For MCQs, confidence computation is straightforward, as only a single token is output (Sec.~\ref{Sec:explorationalgorithm}). However, for OQs, confidence must be aggregated across the entire sequence of tokens. Similar to calibration studies for LLMs~\cite{liu2023litcab}, we investigate various aggregation metrics in Sec.~\ref{Sec:CalibrationSupp}, identifying the geometric average (Eq.\ref{eq:geom_avg}) as the most suitable approach for VLMs. Table~\ref{tab:calibexp} presents the calibration analysis for SOTA VLMs, with all methods showing strong calibration in MCQs. For OQs, Qwen2.5-VL significantly outperforms other approaches. 
The Section~\ref{Sec:CalibrationSupp} of the supplementary material includes deeper calibration analysis, including the corresponding models' calibration plots.

\begin{table}[!tb]
\centering
\caption{Calibration and performance analysis for commonly used VLMs on   GT 1min-length clips in the test split of FALCON-Bench.}
\label{tab:calibexp}
\resizebox{\columnwidth}{!}{%
\begin{tabular}{ccccccccccc}
\toprule
Model & \multicolumn{3}{c}{Visual Information} & \multicolumn{3}{c}{\textbf{MCQs}} & \multicolumn{4}{c}{\textbf{OQs}}  \\
\cmidrule(r){1-1} \cmidrule(r){2-4} \cmidrule(r){5-7} \cmidrule(r){8-11}
Name & \(\Delta t\) & \multicolumn{2}{c}{Frames} & \multicolumn{1}{c}{Acc.} & \multicolumn{2}{c}{Calib.} & \multicolumn{2}{c}{GPT} &  \multicolumn{2}{c}{Calib.} \\
\cmidrule(r){1-1} \cmidrule(r){2-2}  \cmidrule(r){3-4} \cmidrule(r){5-5} \cmidrule(r){6-7} \cmidrule(r){8-9} \cmidrule(r){10-11} 
 & s & \# & Res. & Avg. & ACE \(\downarrow\) & CC \(\uparrow\) & Acc. & Sc. & ACE \(\downarrow\) & CC \(\uparrow\) \\
 \cmidrule(r){1-1} \cmidrule(r){2-2}  \cmidrule(r){3-4} \cmidrule(r){5-5} \cmidrule(r){6-7} \cmidrule(r){8-9} \cmidrule(r){10-11} 
 LLaVA-Video & 60 & 32 & $336\times336$ & \textbf{79.2} & 0.11 & 0.35 & 52.6 & 2.81 & 0.19 & 0.06 \\
 LLaVA-OV & 60 & 32 & $336\times336$ & 72.2 & 0.35 & 0.0 & 39.8 & 2.15 & 0.27 & 5e-3 \\
 Qwen2.5-VL & 60 & 2fps & $336\times616$ & 73.1 & 0.12 & 0.22 & \textbf{59.5} & \textbf{3.24} & 0.29 & 0.28 \\
\bottomrule
\end{tabular}}
\end{table}

\begin{table}[!t]
\centering
\caption{\textcolor{black}{FALCONEye study of different meta-architectures. We vary the LLM and VLM used in FALCONEye-Pro (first row) and evaluate in both MCQs and OQs on FALCON-Bench validation split, including number of LLM tokens ($\#T$) used per question.}}
\label{tab:metaarchitectures_FALCONEye}
\resizebox{\columnwidth}{!}{%
\begin{tabular}{cc ccc cccc}
\toprule
     \multicolumn{2}{c}{\textbf{FALCONEye}} & \multicolumn{3}{c}{\textbf{MCQs}} & \multicolumn{4}{c}{\textbf{OQs}}   \\
  \cmidrule(r){1-2} \cmidrule(r){3-5} \cmidrule(r){6-9}  
  VLM & LLM & $\#T$ & Acc. & Loc. & $\#T$ & Acc. & Sc. & Loc. \\
  \cmidrule(r){1-2} \cmidrule(r){3-5} \cmidrule(r){6-9}  
  Qwen2.5-VL-7B & GPT-4o-m & 8.4K & 59.6 & 27.3 & 7.6K & 46.6 & 2.47 & 26.9 \\ \midrule
   LLaVA-Video-7B & GPT-4o-m  & \textcolor{black}{6.7K} & \textcolor{black}{52.2} & \textcolor{black}{10.9} & 10.9K & 17.5 & 1.13 & 2.71 \\ 
  \textcolor{black}{Qwen2.5-VL-3B} & \textcolor{black}{GPT-4o-m} & \textcolor{black}{11.9K} & \textcolor{black}{57.1} & \textcolor{black}{25.6} & \textcolor{black}{9.4K} & \textcolor{black}{22.8} & \textcolor{black}{1.24} & \textcolor{black}{13.9} \\ 
\midrule
  Qwen2.5-VL-7B & GPT4o & \textcolor{black}{6.5K} & \textcolor{black}{59.7} & \textcolor{black}{33.4} & 8.7K & 48.8 & 2.56 & 23.8 \\ 
   Qwen2.5-VL-7B & GPTo3-m & \textcolor{black}{27.8K} & \textcolor{black}{63.4} & \textcolor{black}{32.3} & 28.8K & 50.4 & 2.63 & 29.3 \\ 
   \textcolor{black}{Qwen2.5-VL-7B} & \textcolor{black}{GPT5} & \textcolor{black}{14.0K} & \textcolor{black}{69.2} & \textcolor{black}{47.2} & \textcolor{black}{18.8K} & \textcolor{black}{47.1} & \textcolor{black}{2.49} & \textcolor{black}{35.5}\\
   \textcolor{black}{Qwen2.5-VL-7B} & \textcolor{black}{Gem2.5-Flash Lite} & \textcolor{black}{21.0K} & \textcolor{black}{56.6} & \textcolor{black}{27.0} & \textcolor{black}{18.4K} & \textcolor{black}{36.1} & \textcolor{black}{1.96} & \textcolor{black}{21.8} \\ 
  \textcolor{black}{Qwen2.5-VL-7B} & \textcolor{black}{Gem2.5-Flash} & \textcolor{black}{28.4K} & \textcolor{black}{69.8} & \textcolor{black}{38.2} & \textcolor{black}{32.0K} & \textcolor{black}{47.3} & \textcolor{black}{2.59} & \textcolor{black}{36.9} \\ 
    \textcolor{black}{Qwen2.5-VL-7B} & \textcolor{black}{Gem2.5-Pro} & \textcolor{black}{12.3K} & \textcolor{black}{69.5} & \textcolor{black}{39.6} & \textcolor{black}{17.2K} & \textcolor{black}{52.3} & \textcolor{black}{2.88} & \textcolor{black}{41.7} \\ 
\bottomrule
\end{tabular}
}
\end{table}

\begin{table}[!t]
\centering
\caption{\textcolor{black}{Study of FALCONEye confidence estimation method.}}
\label{tab:confmethods_FALCONEye}
\resizebox{0.75\columnwidth}{!}{%
\begin{tabular}{c cc ccc}
\toprule
\multicolumn{1}{c}{\textbf{FALCONEye}} & \multicolumn{2}{c}{\textbf{MCQs}} & \multicolumn{3}{c}{\textbf{OQs}}    \\
  \cmidrule(r){1-1} \cmidrule(r){2-3} \cmidrule(r){4-6} 
   Confidence & Acc. & Loc. & Acc. & Sc. & Loc. \\
  \cmidrule(r){1-1} \cmidrule(r){2-3} \cmidrule(r){4-6} 
  No Confidence & 42.0 & 21.7 & 33.4 & 1.84 & 22.0 \\
  VLM Verbal Confidence & 53.4 & 24.3 & 35.4 & 1.87 & 23.0 \\
  LLM Verbal Confidence & 41.2 & 13.5 & 34.8 & 1.82 & 20.6 \\
  \midrule
  Geom.Avg (Ours) & 59.6 & 27.3 & 46.6 & 2.47 & 26.9 \\ 
\bottomrule
\end{tabular}
}
\end{table}

\vspace{-0.5cm}

\paragraph{FALCONEye.}
\label{Sec:FALCONEyeAblation}
\textcolor{black}{We validate FALCONEye's main features by first illustrating its training-free and model agnostic capabilities in Table \ref{tab:metaarchitectures_FALCONEye}, showing that current FALCONEye-Pro performance could be easily increased using a more powerful LLM or even VLM. 
Second, Table \ref{tab:confmethods_FALCONEye} shows  the significance of using a calibrated confidence estimation of the VLM answers inside our exploration algorithm. To achieve this, we compare our confidence estimation mechanism against those used in related works: verbal confidence estimated together with the answer (by the VLM in our case) \cite{zhi2025videoagent2} and verbal confidence estimated separately from the answer by an LLM given the question, answer and video context \cite{wang2024videoagent}.
We also perform a detailed ablation of the different FALCONEye's exploration algorithm stages to show their contribution, including the zoom-in effect. The details of this study are in Sec.~\ref{Sec:ablation-sup} of  supplementary material.}

\section{Conclusions}

We present \textbf{FALCONEye}, a novel video agent that integrates a VLM and an LLM via an efficient exploration algorithm, enabling the answering of single-detail, open-ended questions over long-form videos. We evaluate our meta-architecture using a 7B-size VLM and a cost-efficient LLM (\emph{academic resources}), and demonstrate that it significantly outperforms sota 7B-size VLMs and similar video agents on FALCON-Bench, a new designed benchmark for the Video Answer Search task. Besides, when applied to shorter videos and broader tasks in MLVU benchmark, it surpasses GPT-4o on single-detail tasks while slashing inference cost by roughly an order of magnitude.

\label{Sec:conclusion}

\section{Acknowledgments}
This work was supported by a DGA scholarship and by DGA project T45\_23R, and grants AIA2025-163563-C31, PID2024-159284NB-I00, PID2021-125514NB-I00 and PID2024-158322OB-I00 funded by MCIN/AEI/10.13039/501100011033 and ERDF.

{
    \small
    \bibliographystyle{ieeenat_fullname}
    \bibliography{main}
}
\clearpage
\setcounter{page}{1}
\maketitlesupplementary

\section{FALCON-Bench: more details}
\label{Sec:bench-details}

\subsection{Examples and video source details}\label{subsec:examples}
\begin{figure}[!h]
    \centering
    \includegraphics[width=\linewidth]{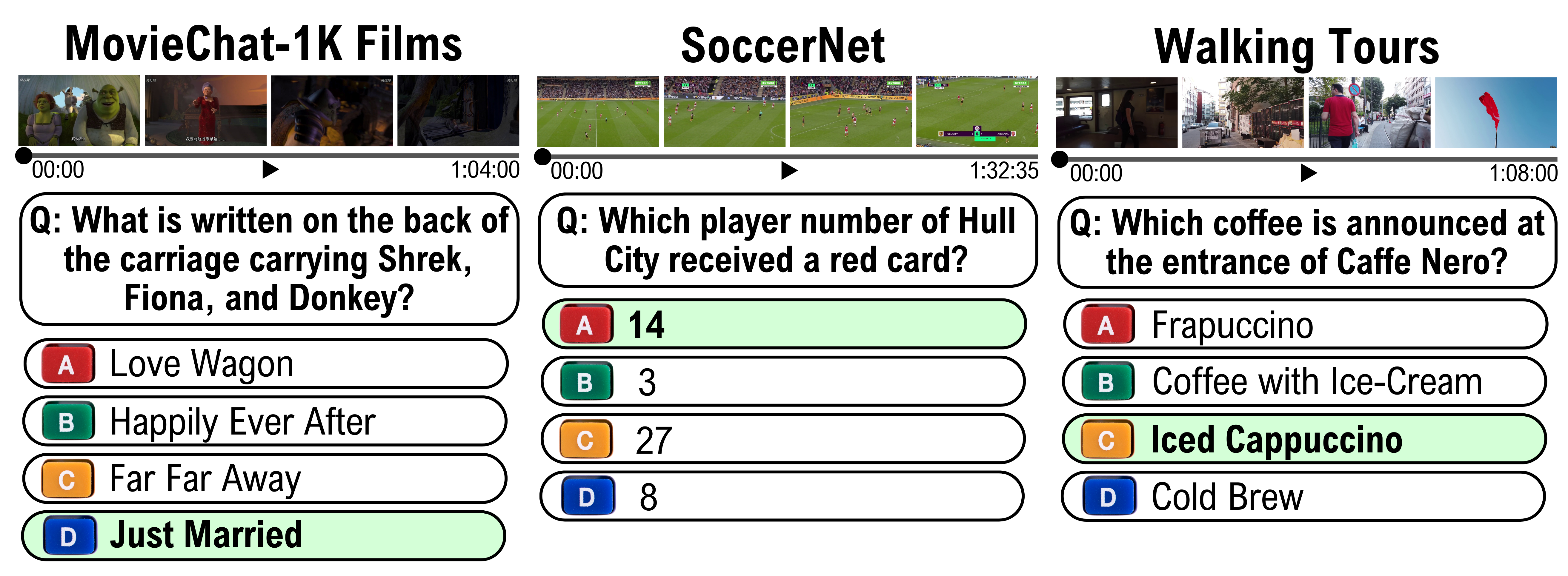}
    \caption{Falcon-Bench question examples for each dataset.}
    \label{fig:FALCON-Bench_examples}
\end{figure}

The videos of our benchmark were sourced from three different public datasets: 

\begin{itemize}
    \item \textbf{S}occerNet \cite{giancola2018soccernet} – 56 videos were selected from this dataset, each averaging 92.4 minutes in duration. These structured soccer match recordings include 389 questions. The mean GT time interval is 61.9 seconds.  
    \item \textbf{M}ovieChat-1K Films \cite{song2024moviechat} – a total of 140 film clips, each 8 minutes long, were selected from the dataset. Clips from the same film were combined to create 24 film segments, with an average duration of 46.4 minutes each. This subset contains 122 questions, with a mean GT time interval of 22.4 seconds.  
    \item Walking \textbf{T}ours Dataset \cite{venkataramanan2023imagenet} – 12 high-resolution videos ($3840\times2160$) were selected from this dataset, averaging 81.3 minutes. These videos capture city tours from an egocentric perspective and include 65 questions. The mean GT time interval is 31.4 seconds.  
\end{itemize}  
Overall, the benchmark comprises 575 questions, covering 4 categories, over 90 videos, with an \textbf{average video duration of 78.9 minutes} and \textbf{answers localized within a GT temporal window of 38.4 seconds}. The dataset is split into a test set (506 questions) and a validation set (70 questions). Figure \ref{fig:FALCON-Bench_examples} shows an example question of our benchmark across each dataset.

\subsection{Question Categories}

To design a benchmark for long VAS tasks, each question should have its answer contained within a single, short clip of the video. Based on this consideration, we defined four question categories:

\begin{itemize} \item \textbf{Text Reading (TR)}: Questions ask about a piece of text that appears at a certain moment in the video. \item \textbf{Visual Observation (VO)}: Questions focus on visual attributes of the items appearing in the video, such as colors, textures, components, or materials. \item \textbf{Time Identification (TI)}: Questions about timestamps on clocks or alarms shown in the video. \item \textbf{Object Identification (OI)}: Questions focus on identifying specific objects within the video. \end{itemize}

\begin{figure}[!h]
    \centering
    \includegraphics[width=0.7\linewidth]{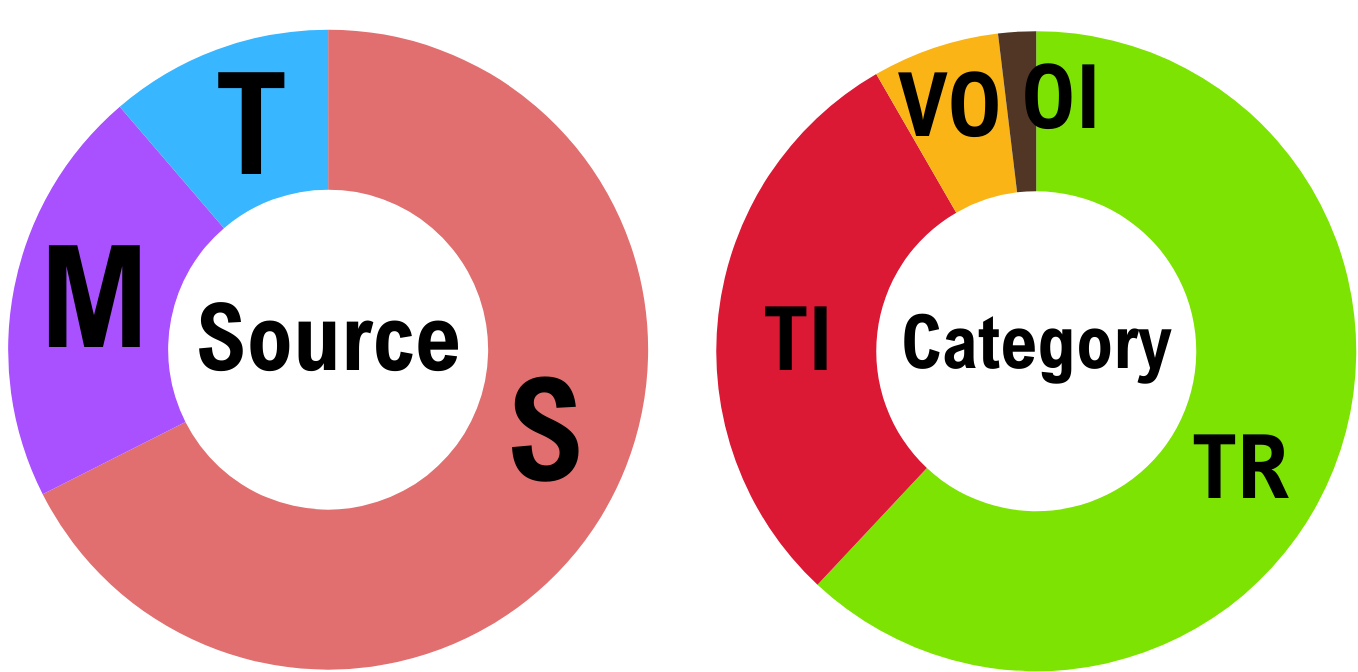}
    \caption{Distribution of questions in Falcon-Bench. The left plot, according to dataset sources: \textbf{M}ovieChat-1k, \textbf{S}occerNet, and Walking\textbf{T}ours. The right plot according to question category: TR, VO, TI, and OI.}
    \label{fig:FALCON-Bench_question_metrics}
\end{figure}

Figure~\ref{fig:FALCON-Bench_question_metrics} provides an overview of the number of questions per video type and per category, illustrating the distribution of tasks across the benchmark.

\subsection{Localization evaluation}\label{Sec:GToU}

\begin{figure}[!htb]
    \centering\includegraphics[width=0.7\linewidth]{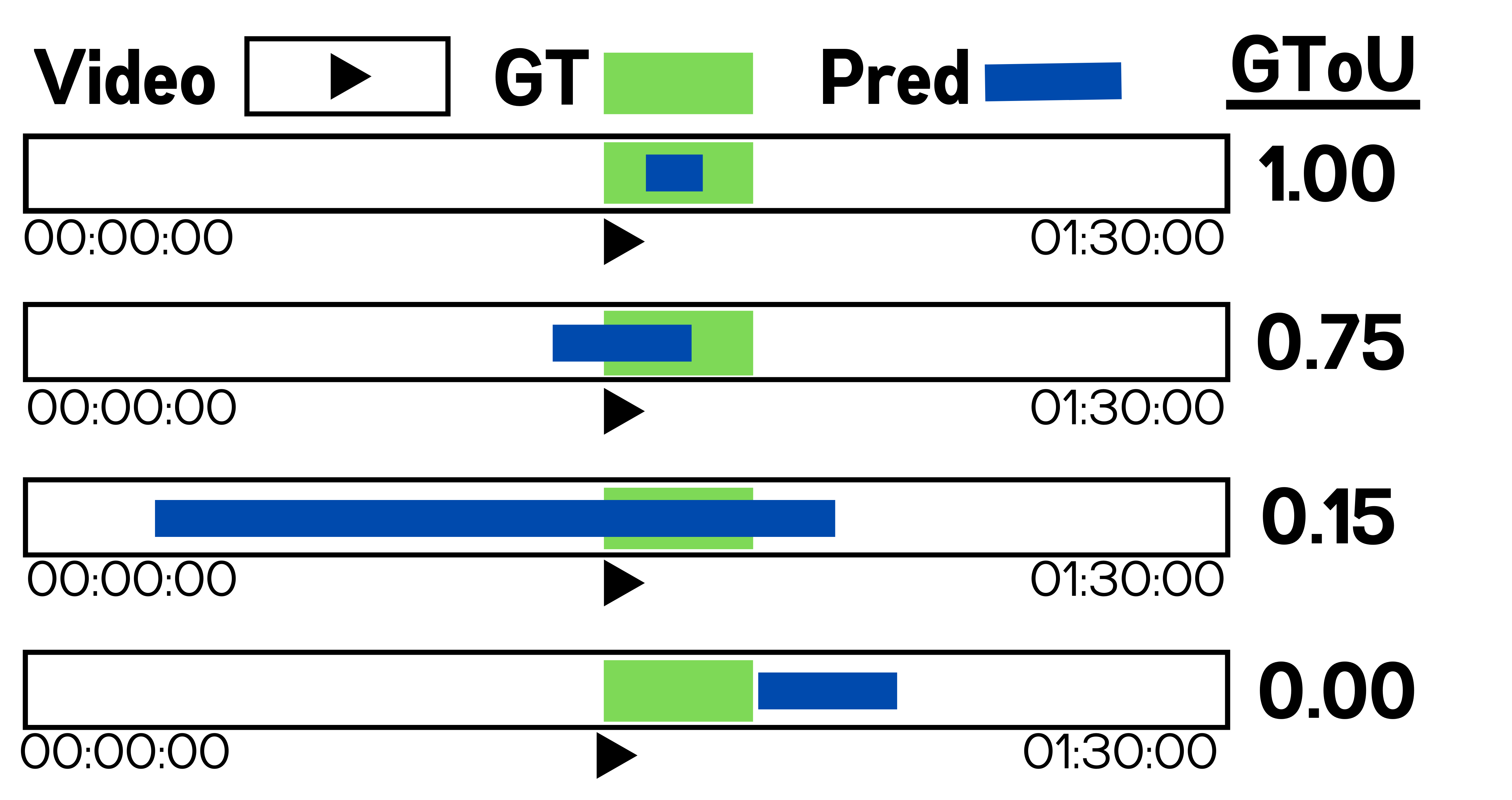}
    \caption{Visualization of  GToU metric designed to measure the clip localization/retrieval task in which the answer is contained.}
    \label{fig:GToUMetric}
\end{figure}

FALCON-Bench requires models to provide an evidence of the answer (short clip) rather than precisely matching the entire clip where the answer is located. To achieve this goal, we leverage The Ground Truth over Union (GToU) metric to compare the predicted and the GT temporal interval. Unlike the commonly used Intersection over Union (IoU) in temporal grounding tasks~\cite{ding2023temporal}, GToU assigns a score of 1.0 if the predicted interval is entirely within the GT interval, regardless of the degree of overlap. In all other cases, GToU behaves identically to IoU (Figure~\ref{fig:GToUMetric}). In mathematical terms,

\begin{equation}
    \textbf{GToU} = \frac{|GT|}{|GT \cup \text{Pred}|} \cdot \mathbbm{1}_{\{|GT \cap \text{Pred}| > 0\}}.
\end{equation}

\subsection{Human experiments}~\label{Sec:HumansExp}
To evaluate human performance on our benchmark, we performed experiments with 10 participants. Each participant answered 10 questions based on 10 different videos (one question per video) and equally spread across the different dataset sources. Each participant answered 5 MCQs and 5 OQs.

\begin{figure*}[!tb]
\centering
\includegraphics[width=0.32\textwidth]{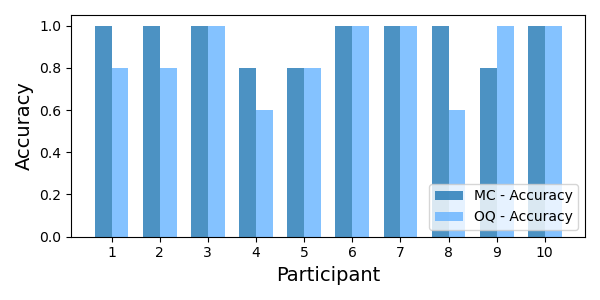}
\includegraphics[width=0.32\textwidth]{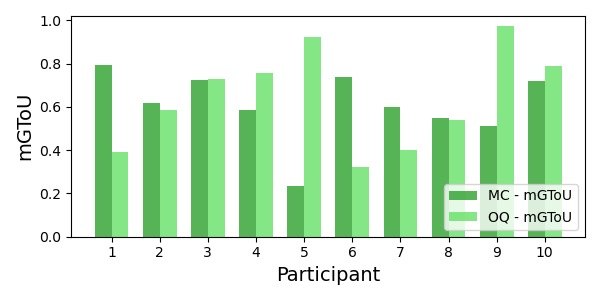}
\includegraphics[width=0.32\textwidth]{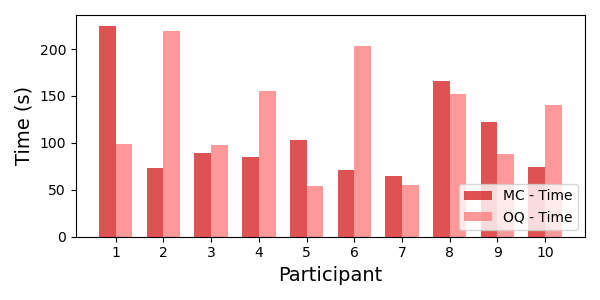}

\caption{Performance metrics across all participants. Figures show accuracy, mean GToU (mGToU), and time spent per question.}
\label{fig:performance_metrics}
\end{figure*}

\paragraph{Methodology} Participants were seated in front a test computer equipped with two monitors, one for watching the video and another for editing a .json file. They were provided with a structured .json file containing the question details.

Participants were required to fill in the \texttt{answer} and \texttt{temporal\_window} fields. About the answer, if the \texttt{options} field is \texttt{null}, the answer was open-ended; otherwise, the answer was a letter in [A, B, C, D]. In the timestamp the participants had to indicate a short video clip where the answer is observed. A supervisor recorded the total time spent for each question, starting when the participant opened the video and stopping when they finished entering both required fields.

\paragraph{Results}

The results of the human experiments were analyzed in terms of both individual performance of each participant and the aggregated performance across all 10 participants, in terms of accuracy, mean GToU (mGToU) and spent time to answer. Figure \ref{fig:performance_metrics} shows individual performance per participant for both MCQs and OQs. Participants 2, 3, 4, and 9 answered correctly all questions. Regarding the spent time, participant 3 was the fastest. Additionally, Table~\ref{tab:human_results} shows the mean accuracy, mGToU, time and score (only for open-ended questions) of the answers across the three video types (MovieChat-1K, SoccerNet, WalkingTours). SoccerNet questions are generally easier to answer correctly, simpler to locate within the video, and quicker to respond to. In contrast, WalkingTours questions are the most challenging overall due to the extended duration and continuous nature of the videos.

\begin{table}[!tb]
\centering
\caption{Mean performance metrics of all participants across the three video types (MovieChat-1K, SoccerNet, WalkingTours).}
\label{tab:human_results}
\resizebox{\columnwidth}{!}{%
\begin{tabular}{cccccccc}
\toprule
\textbf{Dataset} & \multicolumn{2}{c}{\textbf{Accuracy}} & \multicolumn{2}{c}{\textbf{mGToU}} & \multicolumn{2}{c}{\textbf{Time (s)}} & \textbf{Score (0-5)} \\
\cmidrule(r){2-3} \cmidrule(r){4-5} \cmidrule(r){6-7} \cmidrule(r){8-8}
 & MCQs & OQs & MCQs & OQs & MCQs & OQs & OQs \\
\midrule
MovieChat-1k & 100.0 & 84.6 & 63.6 & 60.1 & 118.5 & 120.2 & 4.31 \\
SoccerNet & 100.0 & 95.0 & 79.5 & 89.2 & 55.2 & 67.4 & 4.80 \\
WalkingTours & 76.9 & 76.5 & 76.9 & 58.2 & 173.1 & 200.2 & 4.06 \\
\bottomrule
\end{tabular}}
\end{table}

\section{FALCONEye: more details}
\label{Sec:falcon-details}
This section gives further implementation details of our FALCONEye video agent explained in Sec.~\ref{Sec:method}.

\subsection{Exploration Algorithm pseudo-code}
Algorithm \ref{alg:FALCONEye} shows the pseudo-code of FALCONEye exploration algorithm (explained in Sec. \ref{Sec:explorationalgorithm} and  Fig.~\ref{fig:FALCONEyeExploration}).

% Customizing algorithm appearance for two-column format
\SetAlgoNlRelativeSize{-2}  % Reduce line number size
\SetAlgoNlRelativeSize{-1}  % Make numbering more compact
\RestyleAlgo{algoruled}  % Adds top and bottom rules
\SetAlCapSkip{0.5em}  % Reduce space between caption and algorithm

\begin{algorithm}[!htb]  % 'tb' ensures placement at top or bottom of the column
\small  % Reduce font size for better fit in two-column layout
\caption{FALCONEye Exploration Algorithm}
\label{alg:FALCONEye}

\KwIn{Video $\mathcal{V}$, Question $Q$}
\KwOut{Answer $\mathcal{A}$, Confidence $p(\mathcal{A})$, $Clip$ $v \in \mathcal{V}$}
\textbf{Hyperparams: }{$it_{max}, dur_{t}$} \\

\textbf{\textcolor{gray}{Stage $\circled{1}$ Pre-processing}} \\
$ Clips \gets \text{Segment}(\mathcal{V})$ \\
%\textbf{Step 2: Generate Captions for each initial clip} \\
$\mathcal{C} \gets \text{VLM}(Clips)$ \\
$\mathcal{S} \gets \text{LLM}(\mathcal{C})$ \\

% \textbf{Step 3: Iterative Exploration} \\
$AllAns \gets [\;\;] \;\; ; \;\; it \gets 0 $ \\
\While{$it \leq it_{max}$}{
    \textbf{\textcolor{gray}{Stage \circled{2} Reasoning}} \\
    $Cand\_Clips \gets \text{LLM}(Q, Captions)$  \\
    $Cand\_Captions \gets Cand\_Clips.Captions$ \\
    \While{$dur(Cand\_Clips) \geq dur_{t}$}{
        $Ans \gets [\;\;] $ \\
        \textbf{\textcolor{gray}{Stage \circled{3} Evaluation}} \\ 
        \ForEach{$v_i^{*} \in Cand\_Clips$}{
            $\mathcal{A}_{i}^{*}, p(\mathcal{A}_{i}^{*}) \gets \text{VLM}(Q, v_i^{*})$   \\
            $Ans.append(\{\mathcal{A}_{i}^{*}, p(\mathcal{A}_{i}^{*}), v_{i}^{*}, c_{i}\})$ \\ 
            $it \gets it + 1$ \\
            }
        \textbf{\textcolor{gray}{Stage \circled{4} Decision}} \\ 
        $\mathcal{A}, Promis\_Clips \gets \text{LLM}(Q, Ans)$ \\
         \If{$\mathcal{A} \neq None$}{
         \Return $\mathcal{A}, p(\mathcal{A}), v$ \\
         }
         $AllAns.append(Ans)$ \\
         $Cand\_Clips \gets \text{Segment}(Promis\_Clips)$ \\
         }
   $\mathcal{C}.remove(Cand\_Captions)$
   }

$\mathcal{A} \gets \text{LLM}(Q, AllAns) $ \\
\Return $\mathcal{A}, p(\mathcal{A}), v$ 

\end{algorithm}

\subsection{Prompts}\label{Sec:Prompts}
\begin{figure}[!tb]
\centering\includegraphics[width=0.7\linewidth]{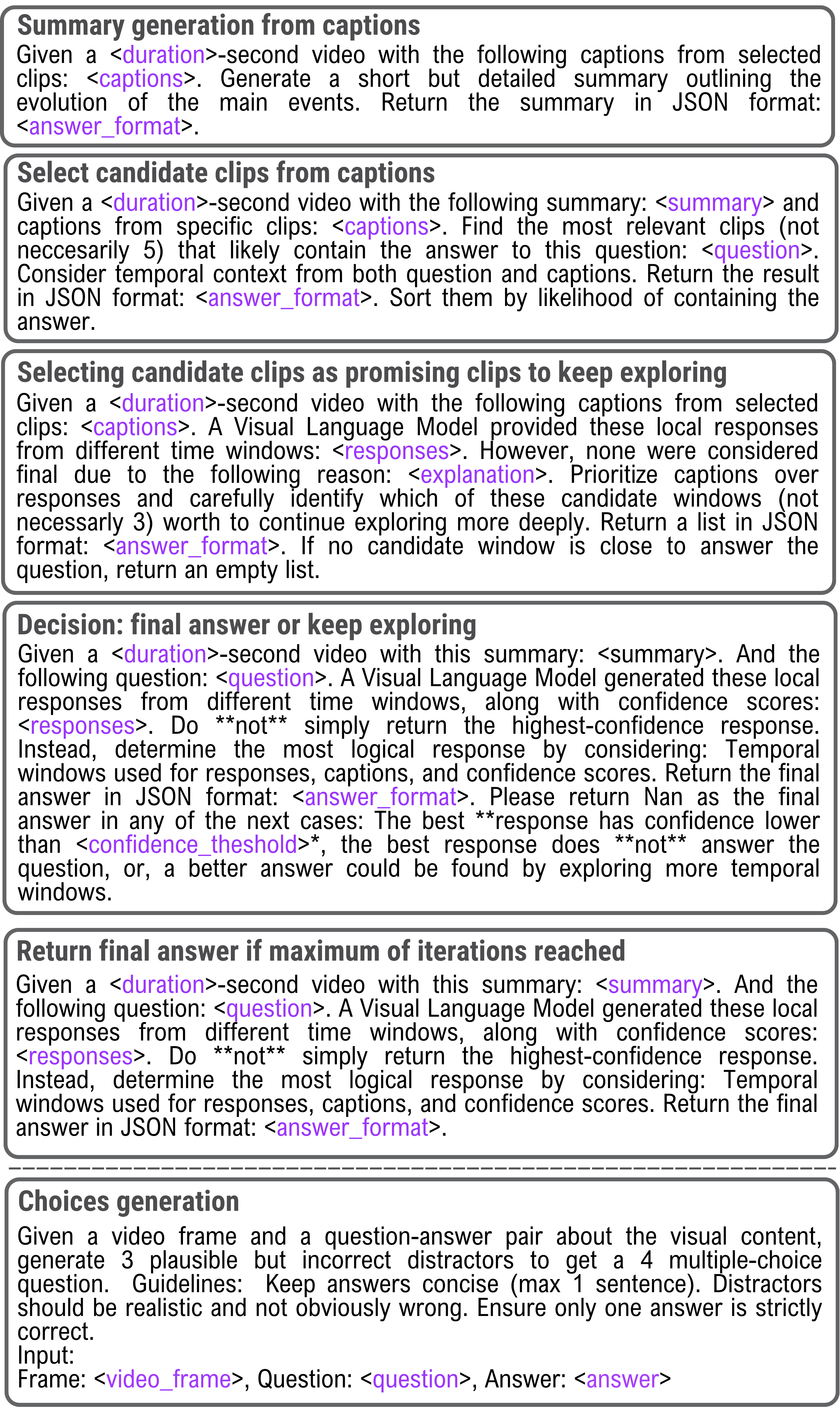}
    \caption{LLM prompts used in FALCONEye algorithm and \textcolor{black}{and FALCON-Bench.}}
    \label{fig:prompts}
\end{figure}

Figure \ref{fig:prompts} shows the prompts sent to the LLM during the different stages of our FALCONEye exploration algorithm. Regarding the stages explained in Sec. \ref{Sec:explorationalgorithm}: \textit{Summary generation} from stage \circled{1} Pre-processing, \textit{Select candidate clips from captions} from stage \circled{2} Reasoning, \textit{Final answer or keep exploring} and \textit{Select candidate clips as promising clips to keep exploring} from stage \circled{4} Decision, and \textit{Return final answer} when the maximum number of evaluated candidate clips is reached.

\section{Additional Experiments}
\subsection{FALCON-Bench}\label{sec:falcon-bench-easy}
Given the poor performance of state-of-the-art VLMs on FALCON-Bench, we further evaluated VLMs under a simplified VAS setup. Specifically, we extracted 1-minute clips centered within the ground truth intervals to ensure that the answers were contained within these shorter segments (Table \ref{tab:gt1min}). Qwen2.5-VL achieved the best accuracy and score in OQs, and LLaVA-Video in MCQs. However, even after reducing the search space from an hour-long video to just one minute, the performance remains relatively low. To address this, we took an additional step and tested the VLMs by providing a single frame that contains the answer, which is always within the ground truth interval defined by FALCON-Bench (Table \ref{tab:gtframe}). These results represent the maximum performance our FALCONEye agent could achieve with each VLM, assuming the ground truth frame is the optimal frame for answering the question. In this evaluation, GPT-4o achieved the highest performance, followed by Qwen2.5-VL.

\begin{table}[!b]
\centering
\caption{Model performance on FALCON-Bench test split with GT 1min-length clips containing the answer. We report average accuracy for MCQs and OQs across MovieChat (M), SoccerNet (S), WalkingTours (T), and overall (Avg.).}
\label{tab:gt1min}
\resizebox{\columnwidth}{!}{%
\begin{tabular}{l c cccccccc}
\toprule
 & MCQs & \multicolumn{8}{c}{\textbf{OQs}}  \\
 \cmidrule(r){2-2} \cmidrule(r){3-10} 
 & Acc. & \multicolumn{4}{c}{Accuracy} & \multicolumn{4}{c}{Score (0-5)} \\
  \cmidrule(r){2-2} \cmidrule(r){3-6} \cmidrule(r){7-10} 
 Model Name &  Avg. & M & S & T & \textbf{Avg.} & M & S & T & \textbf{Avg.} \\
 \midrule
\hline
\rowcolor{gray!10}
\multicolumn{10}{|l|}{\emph{\textbf{Baselines}}} \\
\hline
Full Mark & 100 & 100 & 100 & 100 & 100 & 5.00 & 5.00 & 5.00 & 5.00\\
Random & 25.0 & 0.00 & 0.00 & 0.00 & 0.00 & 0.00 & 0.00 & 0.00 & 0.00 \\
\hline
\rowcolor{gray!10}
\multicolumn{10}{|l|}{\emph{\textbf{Open-Source VLMs with Multi-Image Support}}} \\
\hline
LLaVa-v1.5  & 33.3 & 6.86 & 4.23 & 4.00  & 5.03 & 0.49 & 0.37 & 0.40 & 0.42 \\
LLaVa-v1.6 & 41.9 & 9.80 & 23.4 & 8.00 & 13.7 & 0.58 & 1.33 & 0.58 & 0.83    \\
mPLUG-Owl3 \cite{ye2024mplug} & 57.6 & 39.2 & 21.1 & 26.0 & 28.7 & 2.24 & 1.20 & 1.58 & 1.67 \\
LLaVA-OV \cite{llava-onevision} & 72.2 & 45.0 & 28.5 & 46.0 & 39.8 & 2.46 & 1.57 & 2.42 & 2.15   \\
%LLaMa-3.2M \textcolor{red}{(2)} \cite{llama3_2} & 0.41 & 0.65 & 0.48 & 0.58 &  &  \\
Qwen2.5-VL & 75.9 & 64.7 & 52.2 & 64.0 & 60.3 & 3.49 & 3.08 & 3.26 & 3.27 \\
\hline
\rowcolor{gray!10}
\multicolumn{10}{|l|}{\emph{\textbf{Open-Source VLMs designed for Videos}}} \\
\hline
VideoChat2-HD \cite{li2024mvbench} & 27.2 & 18.6 & 12.1 & 22.0 & 17.5 & 1.24 & 0.67 & 1.34 & 1.08  \\
Video-LLAVA \cite{lin2023video} & 34.0 & 19.6 & 6.49 & 18.0 & 14.6 & 1.20 & 0.46 & 1.14 & 0.93   \\
LLaVA-Video \cite{llava-next} & 79.2 & 61.7 & 44.0 & 52.0 & 52.6 & 3.31 & 2.38 & 2.74 & 2.81 \\
\hline
\rowcolor{gray!10}
\multicolumn{10}{|l|}{\emph{\textbf{Open-Source VLMs specific for long videos}}} \\
\hline
%LongVILA \cite{lin2024vila} &  &  &  &  &  &    \\
MovieChat-OV \cite{song2024moviechat} & 44.5 & 5.88 & 18.3 & 14.0 & 12.7 & 0.49 & 1.02 & 0.82 & 0.77    \\
%LongVU \textcolor{red}{DGX} \cite{shen2024longvu} & 0.81 & 0.44 & 0.84 & 0.56 &  &    \\
Apollo \cite{zohar2024apollo} & 71.3 & 50.9 & 31.3 & 48.0 & 43.4 & 2.77 & 1.79 & 2.62 & 2.39 \\
\hline
\rowcolor{gray!11}
\multicolumn{10}{|l|}{\emph{\textbf{Meta-architectures built from a LLM (GPT4o-mini) and a VLM (Qwen2.5-VL)}}} \\
\hline
\textbf{FALCONEye}\textcolor{black}{-Pro} & 81.7 & 63.7 & 70.0 & 75.0 & 66.7 & 3.39 & 3.51 & 3.56 & 3.49 \\ 
\textbf{FALCONEye}\textcolor{black}{-Flash} & \textbf{81.9} & 66.6 & 69.7 & 70.0 & \textbf{68.8} & 3.54 & 3.62 & 3.62 & \textbf{3.59} \\ 
\bottomrule
\end{tabular}
}
\end{table}

\begin{table}[!tb]
\caption{Model performance on the test set providing the model with GT frames containing the answer. We report average accuracy for MCQs and OQs across MovieChat (M), SoccerNet (S), WalkingTours (T), and overall (Avg.).}
\label{tab:gtframe}
\resizebox{\columnwidth}{!}{%
\begin{tabular}{l c cccccccc}
\toprule
 & MCQs & \multicolumn{8}{c}{\textbf{OQs}}  \\
 \cmidrule(r){2-2} \cmidrule(r){3-10} 
 & Acc. & \multicolumn{4}{c}{Accuracy} & \multicolumn{4}{c}{Score} \\
  \cmidrule(r){2-2} \cmidrule(r){3-6} \cmidrule(r){7-10} 
 Model Name &  Avg. & M & S & T & \textbf{Avg.} & M & S & T & \textbf{Avg.} \\
 \midrule
\hline
\rowcolor{gray!10}
\multicolumn{10}{|l|}{\emph{\textbf{Baselines}}} \\
\hline
Full Mark & 100 & 100 & 100 & 100 & 100 & 5.00 & 5.00 & 5.00 & 5.00\\
Random & 25.0 & 0.00 & 0.00 & 0.00 & 0.00 & 0.00 & 0.00 & 0.00 & 0.00 \\
% Human &  &  &  &  &  &  \\
\hline
\rowcolor{gray!10}
\multicolumn{10}{|l|}{\emph{\textbf{Proprietary Long-Context LLMs}}} \\
\hline
%Gemini-1.5-Pro &  &  &  &  &  &   \\
GPT-4o-mini (LR) & 75.3 & 61.7 & 28.5 & 42.0 & 44.0 & 3.18 & 1.52 & 2.28 & 2.32 \\
GPT-4o-mini (HR) & 88.4 & 70.5 & 57.9 & 70.0 & 66.1 & 3.58 & 3.21 & 3.78 & 3.52  \\
GPT-4o (LR)   & 83.7 & 78.0 & 47.7 & 62.0 & 52.5 & 3.94 & 2.63 & 3.32 & 3.29 \\
GPT-4o (HR) & \textbf{94.1} & 80.3 & 66.9 & 72.0 & \textbf{73.0} & 4.06 & 3.67 & 3.82 & \textbf{3.85} \\
\hline
\rowcolor{gray!10}
\multicolumn{10}{|l|}{\emph{\textbf{Open-Source VLMs for images}}} \\
\hline
LLaVa-v1.5  & 56.7 & 40.1 & 24.5 & 34.0 & 32.8 & 2.34 & 1.38 & 1.96 & 1.89  \\
LlaVa-v1.6  & 78.5 & 53.9 & 48.5 & 62.0 & 54.8 & 3.08 & 2.51 & 3.42 & 3.24     \\
mPLUG-Owl3 \cite{ye2024mplug} & 82.9 & 59.8 & 53.1 & 58.0 & 65.2 & 3.25 & 2.89 & 3.18 & 3.10 \\
LLaVA-OV \cite{llava-onevision} & 84.4 & 40.1 & 10.4 & 54.0 & 34.8 & 2.02 & 0.53 & 2.80 & 1.78  \\
Qwen2.5-VL \cite{} & 84.1 & 71.5 & 62.9 & 78.0 & \underline{70.8} & 3.80 & 3.33 & 4.02 & \underline{3.72} \\
LLaMa3.2M \cite{llama3_2} & \underline{85.9} & 69.6 & 64.9 & 70.0 & 68.1 & 3.68 & 3.27 & 3.70 & 3.55 \\
\bottomrule

\end{tabular}
}
\end{table}

\subsection{Ablation study of exploration algorithm stages}\label{Sec:ablation-sup} 

We validate FALCONEye's exploration algorithm with an ablation study of its different stages, explained in Sec. \ref{Sec:explorationalgorithm}, shown in Table \ref{tab:algorithm_FALCONEye}. It is noted that each stage contributes significantly to achieving the FALCONEye's superior performance compared to the baselines.

\begin{table}[!b]
\centering
\caption{FALCONEye ablation study of the exploration algorithm. We compare the time and performance gain that each of the four stages of our exploration algorithm brings (\textbf{\textcircled{1} Pre-processing}, \textbf{\textcircled{2} Reasoning}, \textbf{\textcircled{3} Evaluation}, and \textbf{\textcircled{4} Decision}, as detailed in Sec. \ref{Sec:method}). To validate them, we first measure performance when giving the whole video to the VLM and the captions extracted in Stage \textcircled{1} to the LLM. Lately, we validate the stages adding them sequentially and comparing the reasoning stages \textcircled{2} and \textcircled{4} vs random guess.}
\label{tab:algorithm_FALCONEye}
\resizebox{\columnwidth}{!}{
\begin{tabular}{c ccc cccc}
\toprule
     \multicolumn{1}{c}{\textbf{FALCONEye}} & \multicolumn{3}{c}{\textbf{MCQs}} & \multicolumn{4}{c}{\textbf{OQs}}   \\
  \cmidrule(r){1-1} \cmidrule(r){2-4} \cmidrule(r){5-8}  
  Exploration Algorithm & s & Acc. & Loc. & s & Acc. & Sc. & Loc. \\
  \cmidrule(r){1-1} \cmidrule(r){2-4} \cmidrule(r){5-8}
  Video $\rightarrow$ \text{VLM} & 68.9 & 23.4 & 0.00 & 70.1 & 11.3 & 0.71 & 0.00 \\
  Captions from $\textcircled{1} \rightarrow$ \text{LLM} & 123.4 & 39.9 & 17.8 & 125.0 & 13.8 & 0.89 & 16.2  \\ 
  $\textcircled{1}$+Random+$\textcircled{3}$ & 174.9 & 26.1 & 3.16 & 165.2 & 9.92 & 0.60 & 4.06  \\ 
  $\textcircled{1}$+$\textcircled{2}$+$\textcircled{3}$ & 181.2 & 51.9 & 22.1 & 171.6 & 34.8 & 1.90 & 21.2 \\ $\textcircled{1}$+$\textcircled{2}$+$\textcircled{3}$+Random & 383.2 & 54.8 & 26.8 & 290.8 & 38.9 & 2.01 & 21.9  \\ 
  \midrule
FALCONEYE:\textcircled{1}+\textcircled{2}+\textcircled{3}+\textcircled{4} & 348.7 & 59.6 & 27.3 & 229.2 & 46.6 & 2.47 & 26.9 \\ 
\bottomrule
\end{tabular}
}
\end{table}
\label{Sec:zoom-in} 
We also perform a more detailed study of the influence of the zoom-in effect incorporated in our approach. Qwen2.5-VL can process images of any resolution, dynamically converting them into a variable number of visual tokens. This creates a trade-off between number of frames and frame resolution when processing videos within the context window size limit. This trade-off is analyzed in the validation split of FALCON-Bench by varying the clip length (Table~\ref{tab:zoom-in-exp}). The best configuration per clip-level category is selected for FALCONEye.

\begin{table}[!tb]
\centering
\caption{Qwen2.5-VL performance comparison when varying the GT clip length of FALCON-Bench validation split.}
\label{tab:zoom-in-exp}
\resizebox{\columnwidth}{!}{%
\begin{tabular}{ccc cccc cccccccc}
\toprule
\multicolumn{3}{c}{Visual Information} & \multicolumn{4}{c}{\textbf{MCQs}} & \multicolumn{7}{c}{\textbf{OQs}}  \\
\cmidrule(r){1-3} \cmidrule(r){4-7} \cmidrule(r){8-15}
$\Delta t$ & \multicolumn{2}{c}{Frames} & \multicolumn{4}{c}{Accuracy} & \multicolumn{4}{c}{Accuracy} &  \multicolumn{4}{c}{Score} \\
\cmidrule(r){1-1} \cmidrule(r){2-3}  \cmidrule(r){4-7} \cmidrule(r){8-11} \cmidrule(r){12-15} 
 s & \# & Res. &  M & S & T & Avg. & M & S & T & Avg. & M & S & T & Avg. \\
\cmidrule(r){1-1} \cmidrule(r){2-3}  \cmidrule(r){4-7}  \cmidrule(r){8-11} \cmidrule(r){12-15} 
 60 & 2fps & $336\times616$ & 85.0 & 54.3 & 73.3 & 70.9 &  60.0 & 57.1 & 66.6 & 61.3 & 3.15 & 3.23 & 3.46 & 3.28\\
  60 & 64 & $476\times840$   & 80.0 & 62.8 & 73.3 & 72.1  & 70.0 & 57.1 & 53.3 & 60.2 & 3.60 & 3.46 & 2.93 & 3.38\\
 \rowcolor{gray!10} 60 & 32 & $672\times1204$  & 85.0 & 71.4 & 86.7 & \textbf{81.0} & 60.0 & 60.0 & 66.7 & \textbf{62.2} & 3.15 & 3.54 & 3.53 & \textbf{3.43}\\
 \midrule
 \rowcolor{gray!10} 5 & 2fps & $824\times1462$ & 85.0 & 74.3 & 80.0 & 79.8 &  70.0 & 62.9 & 73.3 & \textbf{68.7} & 3.70 & 3.71 & 3.80 & \textbf{3.73}\\
 5 & 5 & $824\times1462$  & 85.0 & 80.0 & 93.3 & \textbf{86.1} & 55.0 & 65.7 & 66.7 & 62.5 & 2.95 & 3.69 & 3.53 & 3.39\\
\bottomrule
\end{tabular}}
\end{table}

\subsection{VLMs Calibration}\label{Sec:CalibrationSupp}
To measure VLM calibration, we adopt the Reliability Diagrams~\cite{degroot1983comparison}. These diagrams group all the predictions in bins according to their confidence and measure the gap between confidence and accuracy per each bin. Specifically, we split the $N$ predictions in $M=10$ bins according to their confidence and, for each bin $B_{m}$, we compute its count $N_{m}$ average confidence $C_{m}$ and its average accuracy $A_{m}$. From these diagrams, we may compute the Average Calibration Error (ACE) as, 
\begin{equation}
    \text{ACE} = \frac{1}{M^{+}}\sum_{m=1}^{M}|C_{m}-A_{m}|, \; \;\text{MCE} = \max_{m}|C_{m}-A_{m}|
\end{equation}
where $M^{+}$ is the number of non-empty bins~\cite{neumann2018relaxed}. We compute Calibration Count (CC), which quantifies the percentage of predictions above a defined confidence threshold, weighted by their accuracy calibration error (1-CE). For example, the Calibration Count at threshold 0.9 is computed as,
\begin{equation}
    {\text{CC}@0.9 = \frac{N_{10}}{N}\left(1-|C_{10}-A_{10}|\right)}.
\end{equation}

\begin{figure}[!tb]
    \centering
    \includegraphics[width=\columnwidth]{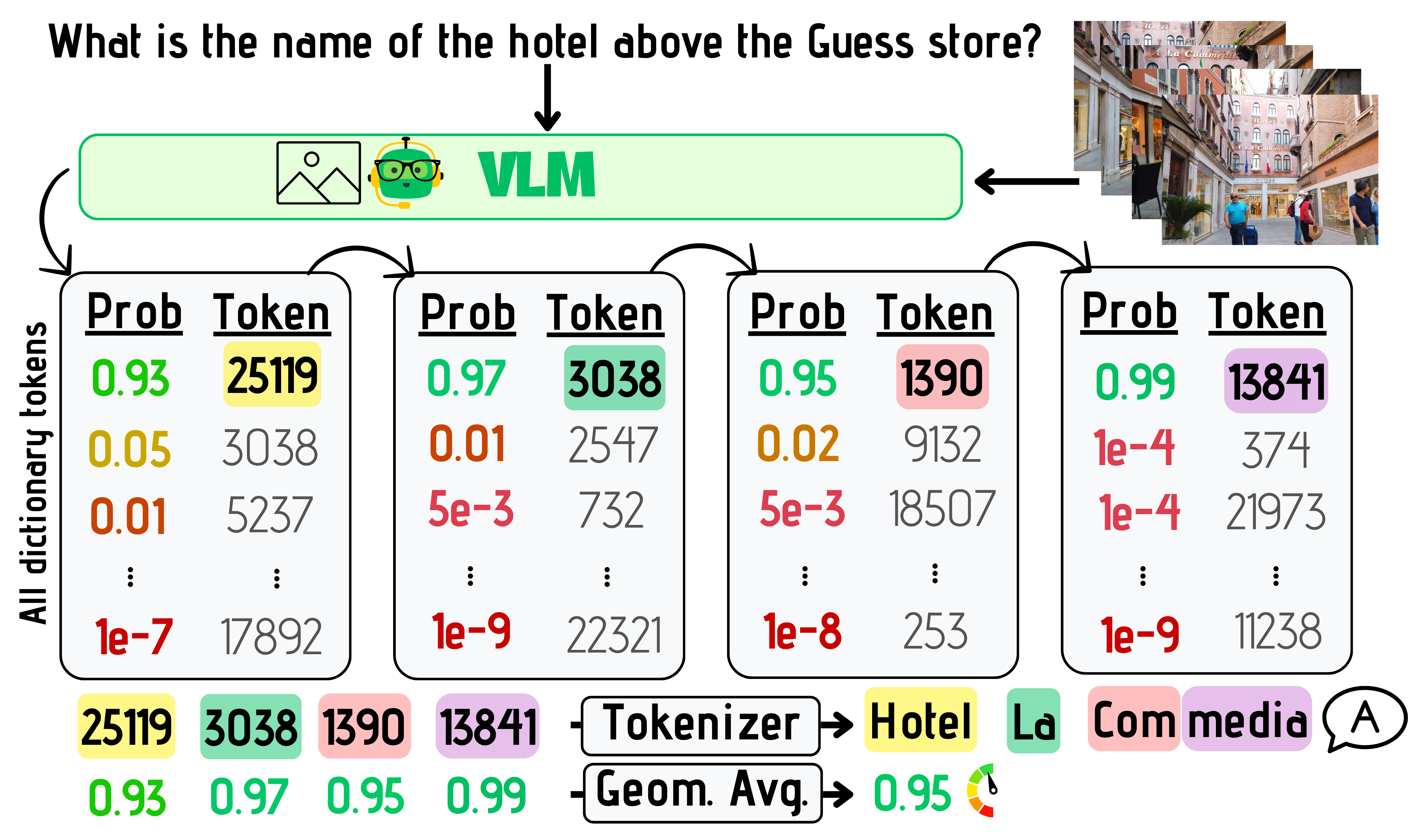}
    \caption{Given a question and a set of frames, \textbf{FALCONEye}  \textbf{leverages} the \textbf{answer} outputted by the VLM but, and its \textbf{confidence} (geometric average through all output tokens probabilities).}
    \label{fig:probsFALCON}
\end{figure}

\paragraph{Confidence metrics-.} 
As detailed in Sec.\ref{Sec:explorationalgorithm}, estimating answer confidence in OQs requires a metric to average token probabilities along the response (see Figure~\ref{fig:probsFALCON}).

\begin{figure*}[!t]
    \centering
\footnotesize
\begin{tabular}{cc|cc}
% FIRST ROW
        \includegraphics[width=0.23\textwidth]{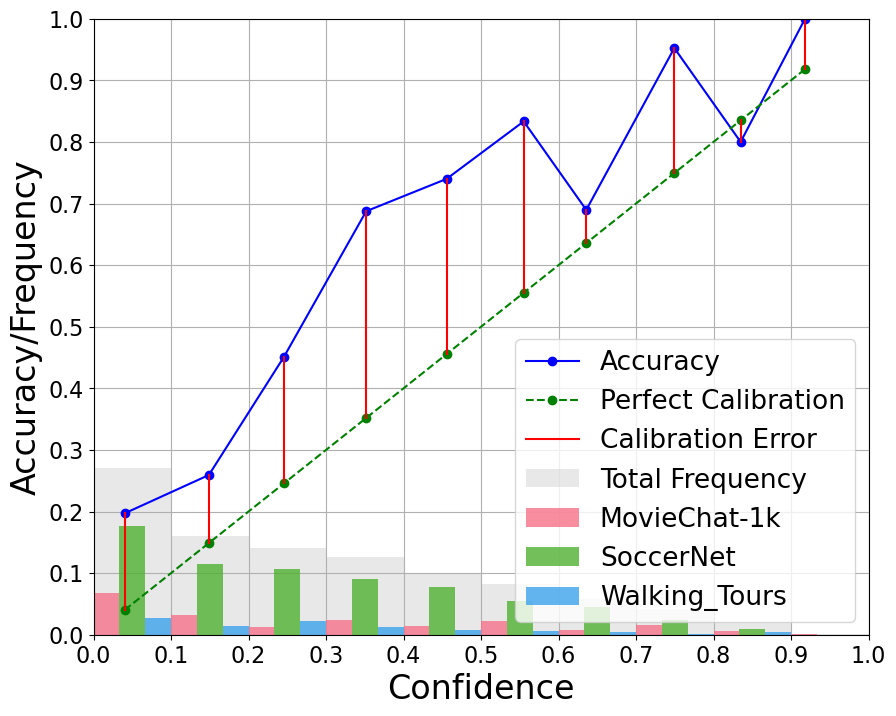} & % Replace with your image file
        %\caption{LLaVA-Video, Likelihood.}
        %\label{fig:calib1}

        \includegraphics[width=0.23\textwidth]{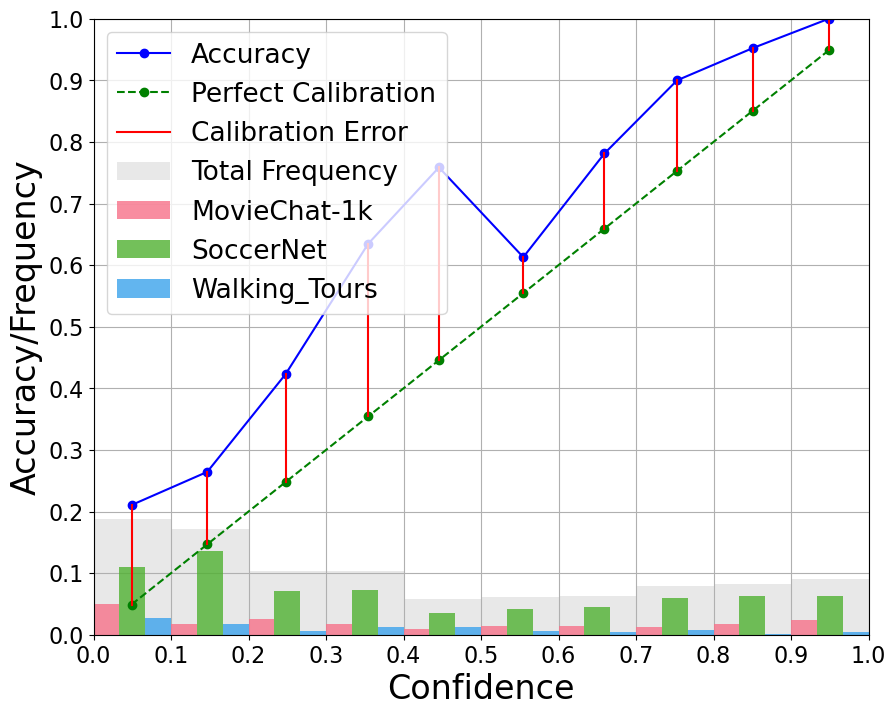} & % Replace with your image file
        %\caption{Qwen2.5-VL-2fps, Likelihood.}
        %\label{fig:calib2}
        \includegraphics[width=0.23\textwidth]{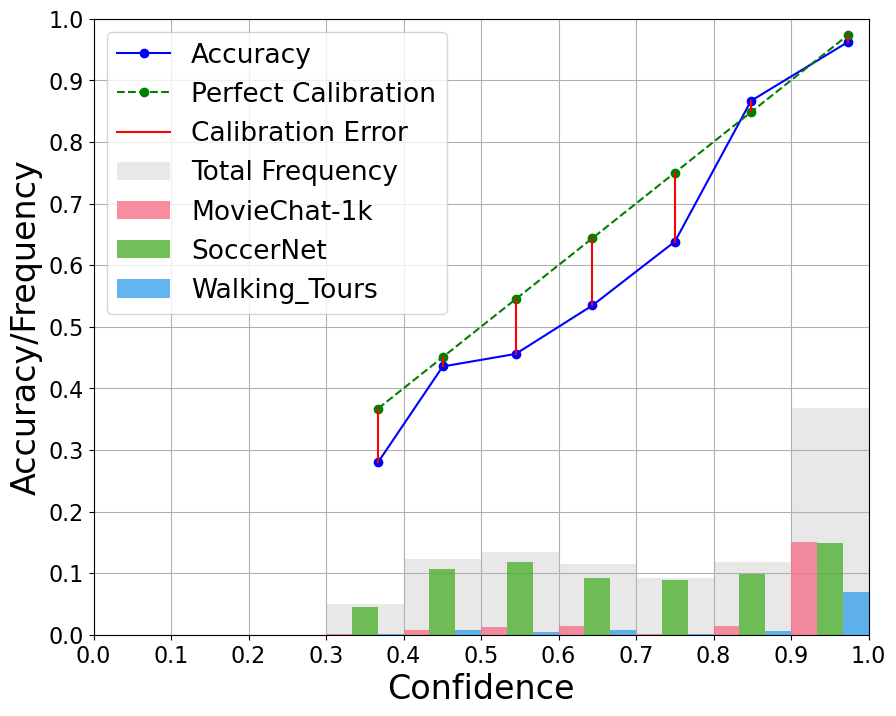} & % Replace with your image file
        %\caption{LLaVA-Video, w/ Options.}
        %\label{fig:calib1}
        \includegraphics[width=0.23\textwidth]{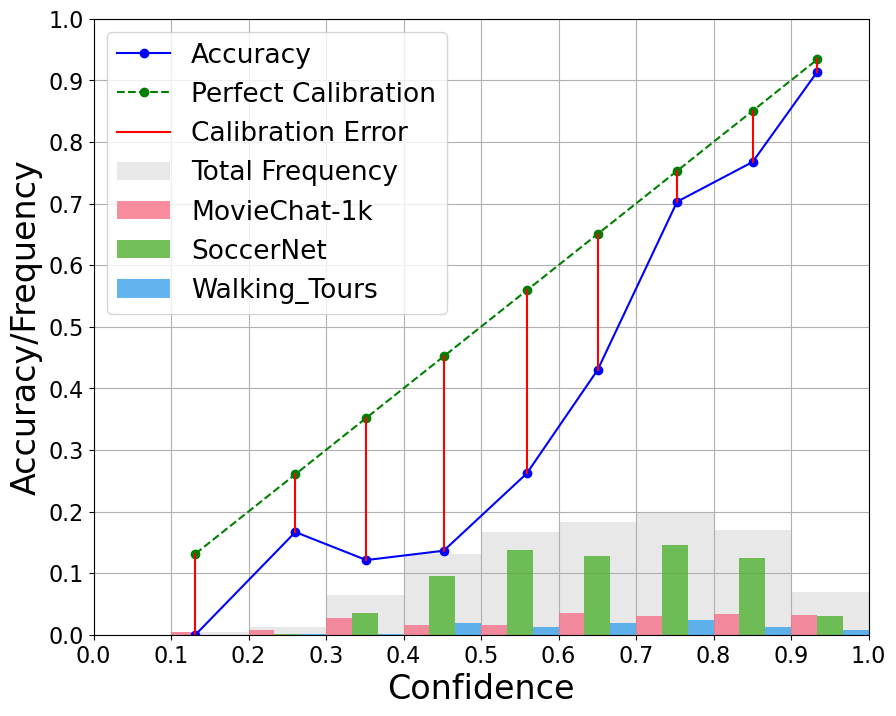}\\ % Replace with your image file
        %\caption{LLaVA-Video, w/o Options.}
        %\label{fig:calib2}
        (a) LLaVA-Video, Likelihood. & 
        (b) Qwen2.5-VL-2fps, Likelihood. &
        (g) LLaVA-Video, MCQs &
        (h) LLaVA-Video, OQs\\

    % % Second row
        \includegraphics[width=0.23\textwidth]{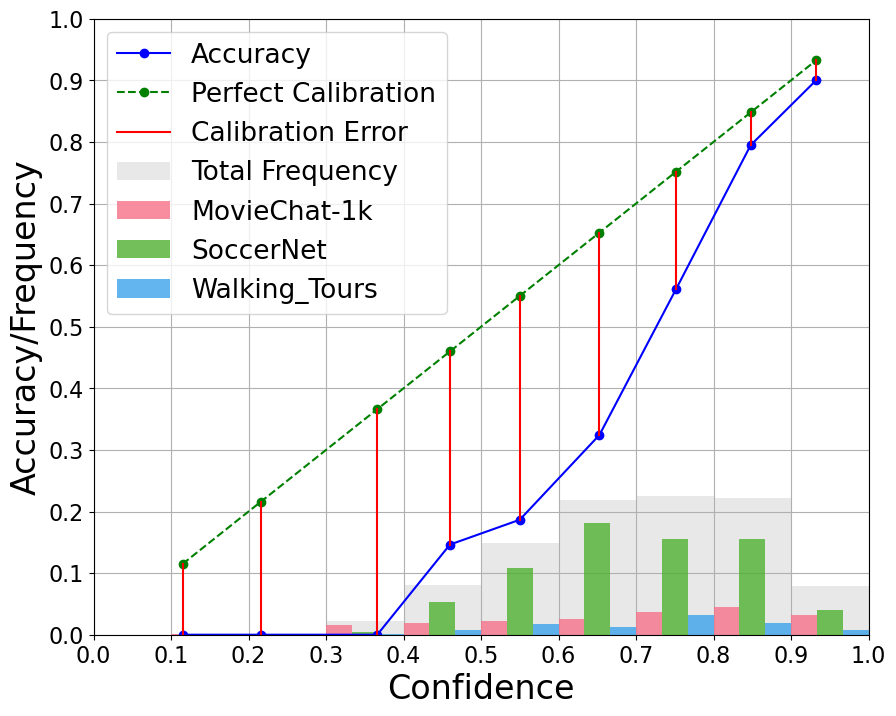} &% Replace with your image file
%        \caption{LLaVA-Video, Average.}
%        \label{fig:calib3}
        \includegraphics[width=0.23\textwidth]{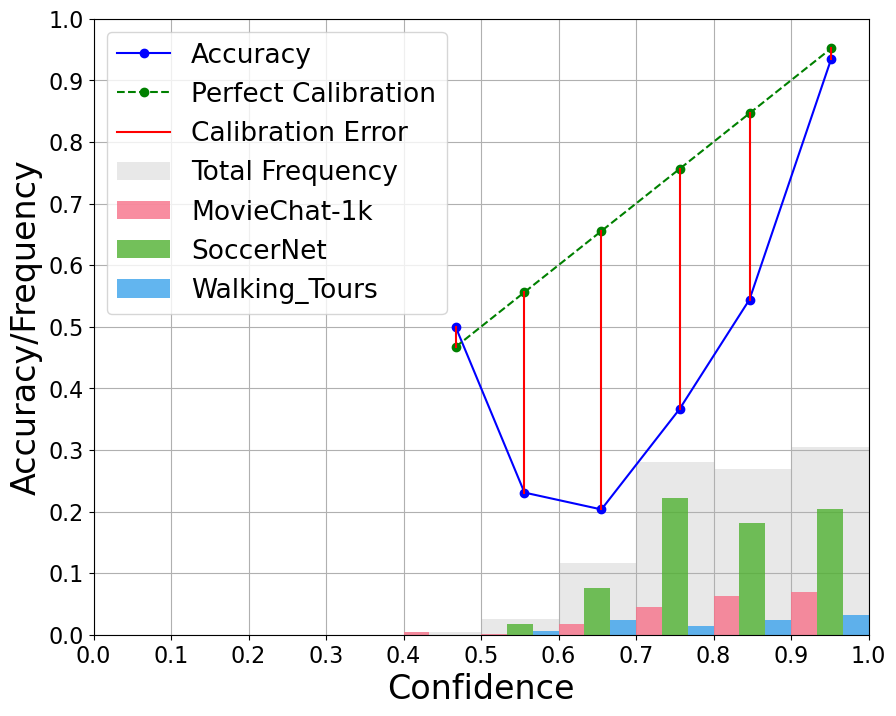} &% Replace with your image file
%        \caption{Qwen2.5-VL-2fps, Average.}
%        \label{fig:calib4}
        \includegraphics[width=0.23\textwidth]{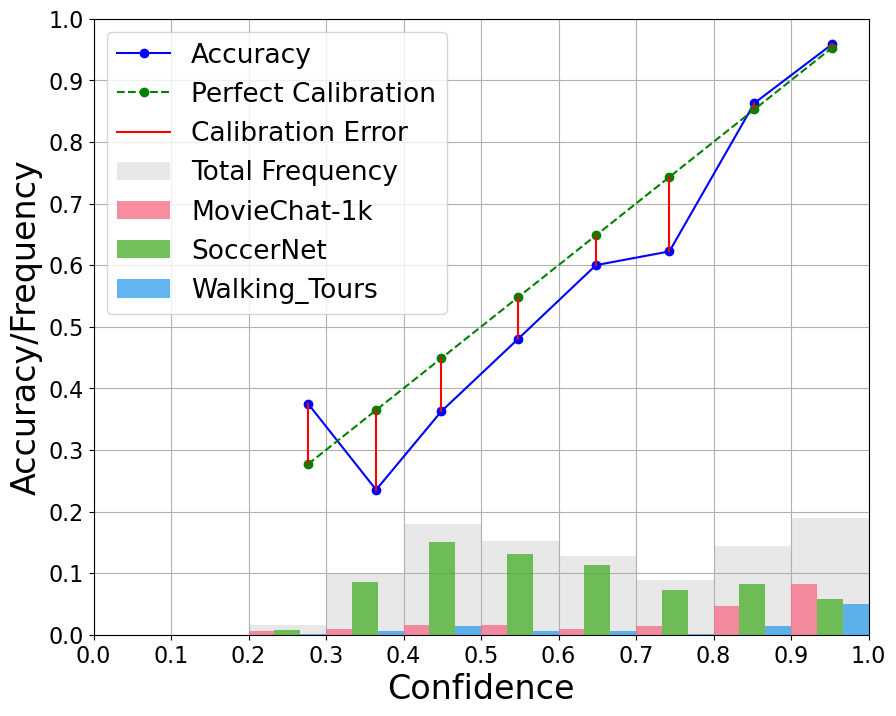} &% Replace with your image file
%        \caption{LLaVA-Video, Geom. avg.}
%        \label{fig:calib3}
        \includegraphics[width=0.23\textwidth]{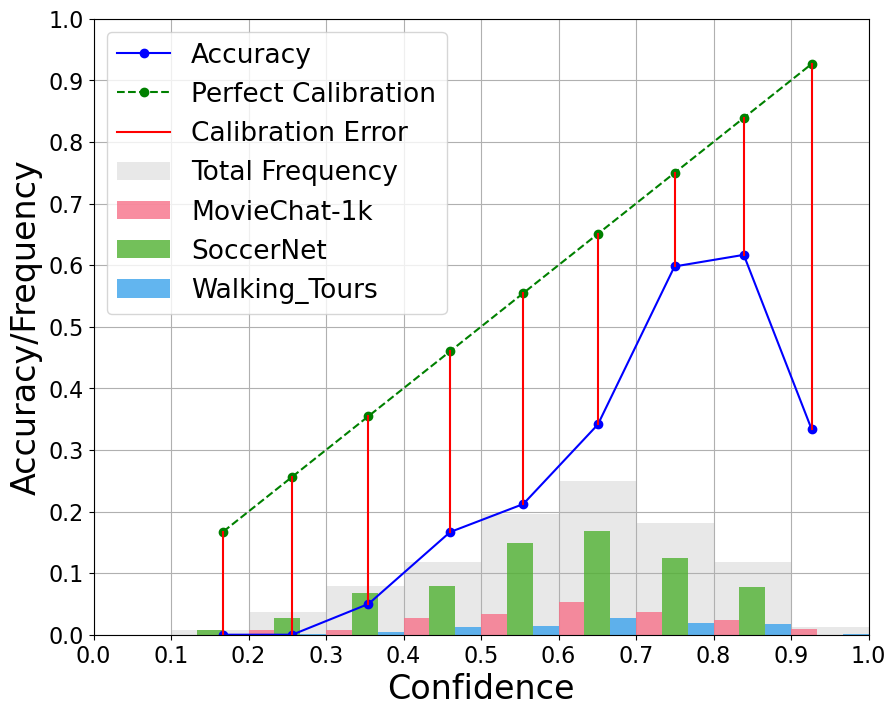}\\ % Replace with your image file
%        \caption{Qwen2.5-VL-2fps, Geom. avg.}
%        \label{fig:calib4}
        (c) LLaVA-Video, Average. &
        (d) Qwen2.5-VL-2fps, Average. &
        (i) LLaVA-OV, MCQs &
        (j) LLaVA-OV, OQs\\

    % Third row
    
        \includegraphics[width=0.23\textwidth]{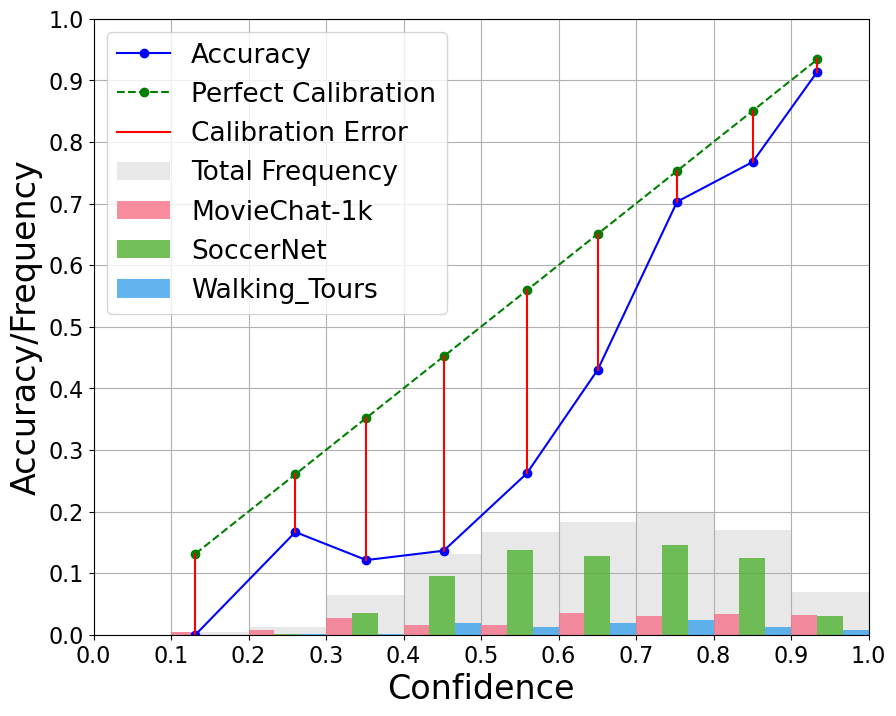} &% Replace with your image file
%        \caption{LLaVA-OV, w/ Options.}
%        \label{fig:calib3}
        \includegraphics[width=0.23\textwidth]{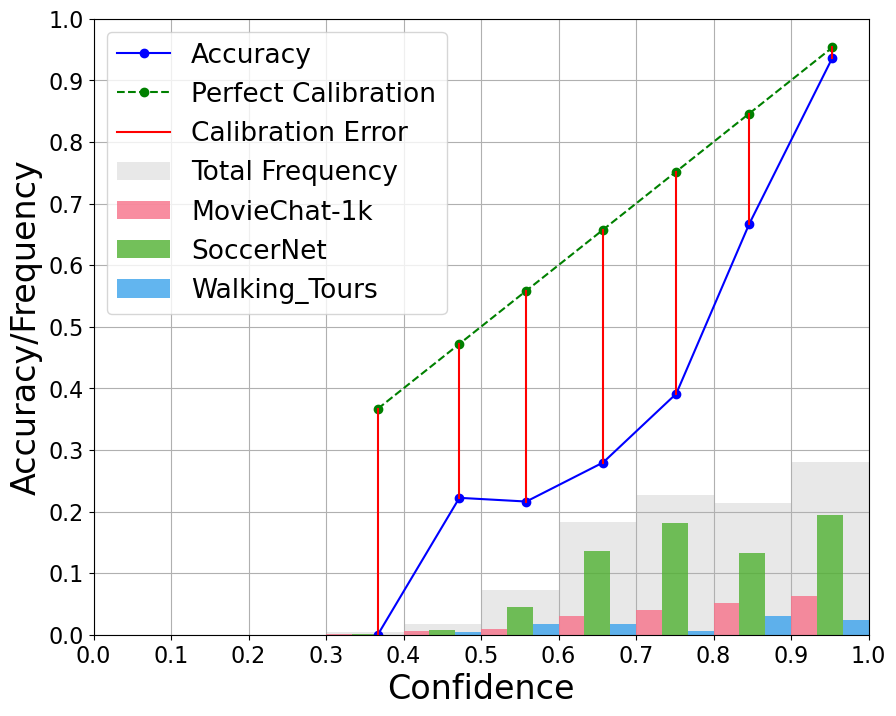} &% Replace with your image file
%        \caption{LLaVA-OV, w/o Options.}
%        \label{fig:calib4}
        \includegraphics[width=0.23\textwidth]{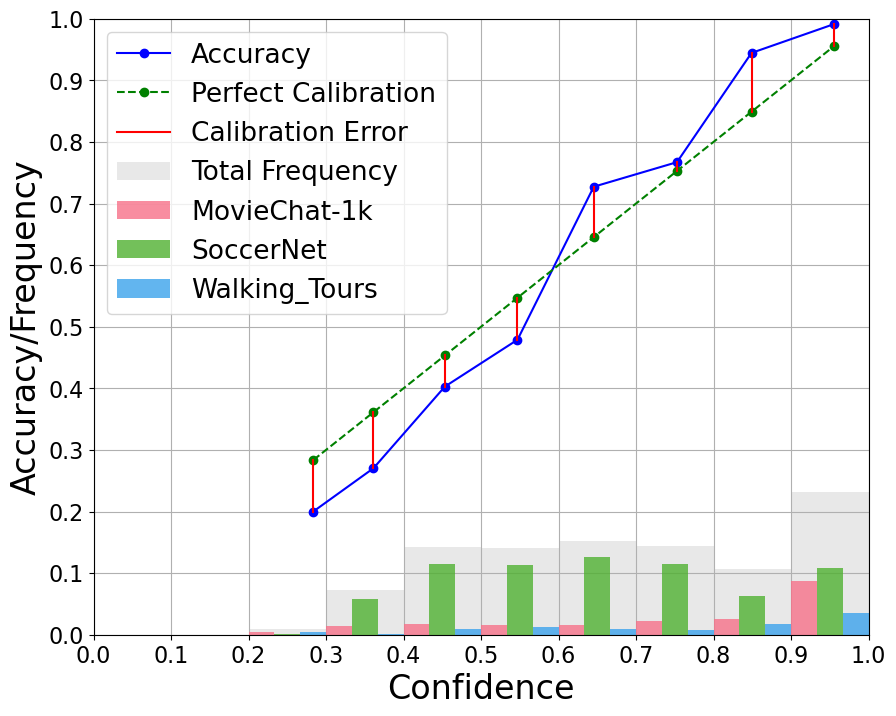} &% Replace with your image file
%        \caption{Qwen2.5-VL-2fps, w/ Options}
%        \label{fig:calib3}
        \includegraphics[width=0.23\textwidth]{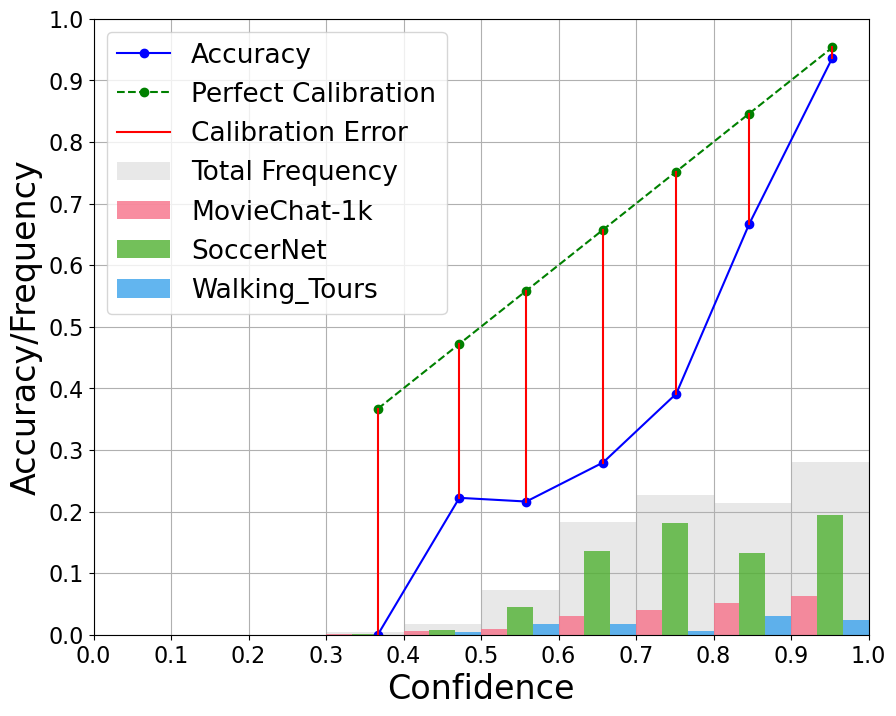} \\% Replace with your image file
%        \caption{Qwen2.5-VL-2fps, w/o Opt.}
%        \label{fig:calib4}
        (e) LLaVA-Video, Geom. avg &
        (f) Qwen2.5-VL-2fps, Geom. avg. &
        (k) Qwen2.5-VL-2fps, MCQs &
        (l) Qwen2.5-VL-2fps, OQs\\
\end{tabular}    
    \caption{(a-f) Calibration plots for different probabilities aggregation metrics with the GT 1min clips of the FALCON-Bench test split, and (g-l) calibration plots when testing the VLMs with the GT 1min-length clips of the FALCON-Bench test split.}
    \label{fig:calibrationmodels}
\end{figure*}

We evaluated various confidence aggregation methods, specifically likelihood, average, and geometric average, as done in similar calibration studies with LLMs~\cite{liu2023litcab}. Looking first at the reliability diagrams (Figure \ref{fig:calibrationmodels} (a-f)), we discard likelihood as a suitable confidence metric. The distribution of confidence values is highly spread out, with many answers assigned extremely low confidence. This occurs because likelihood multiplies the probability of all tokens, thus longer answers tend to receive lower confidence scores, making it unreliable for calibration.

Comparing the average and geometric average, the reliability diagrams do not show major differences between them. However, the calibration metrics in Table~\ref{tab:calibexp_supp} indicate that \textbf{geometric average} results in lower Brier Score (BS), MCE, and ACE for both LLaVA-Video and Qwen2.5-VL, suggesting slightly better calibration performance.

\begin{table}[!tb]
\centering
\caption{Calibration metrics comparison with GT 1min-length clips in the open-ended (OQs) questions test split of FALCON-Bench. Lower values for BS, MCE and ACE, are better.}
\label{tab:calibexp_supp}
\resizebox{\columnwidth}{!}{%
\begin{tabular}{ccc ccc ccc ccc}
\toprule
Model Name & $\Delta t$ & F & \multicolumn{3}{c}{Likelihood} & \multicolumn{3}{c}{Average}  & \multicolumn{3}{c}{Geom. Average} \\
 \cmidrule(r){1-1} \cmidrule(r){2-3}  \cmidrule(r){4-6} \cmidrule(r){7-9} \cmidrule(r){10-12}
 & s & \# & BS & MCE & ACE & BS & MCE & ACE & BS & MCE & ACE  \\
 \cmidrule(r){1-1} \cmidrule(r){2-3}  \cmidrule(r){4-6} \cmidrule(r){7-9} \cmidrule(r){10-12}
 LLaVA-Video & 60 & 32 & 0.23 & 0.34 & 0.17 & 0.24 & 0.37 & 0.22 & 0.22 & 0.31 & 0.16\\
 LLaVA-OV & 60 & 32 & 0.26 & 0.92 & 0.41 & 0.28 & 0.39 & 0.29 & 0.26 & 0.59 & 0.29\\
 %Qwen2.5-VL & 60 & 32 & 651\times1178 & 78.8 & 0.10 & 0.25 & 58.7 & 3.23 & 0.34 & 0.31\\
 Qwen2.5-VL & 60 & 2fps & 0.19 & 0.31 & 0.15 & 0.26 & 0.45 & 0.25 & 0.24 & 0.38 & 0.27\\
\bottomrule
\end{tabular}}
\end{table}

\begin{figure*}[!tb]
    \centering
    \includegraphics[width=\linewidth]{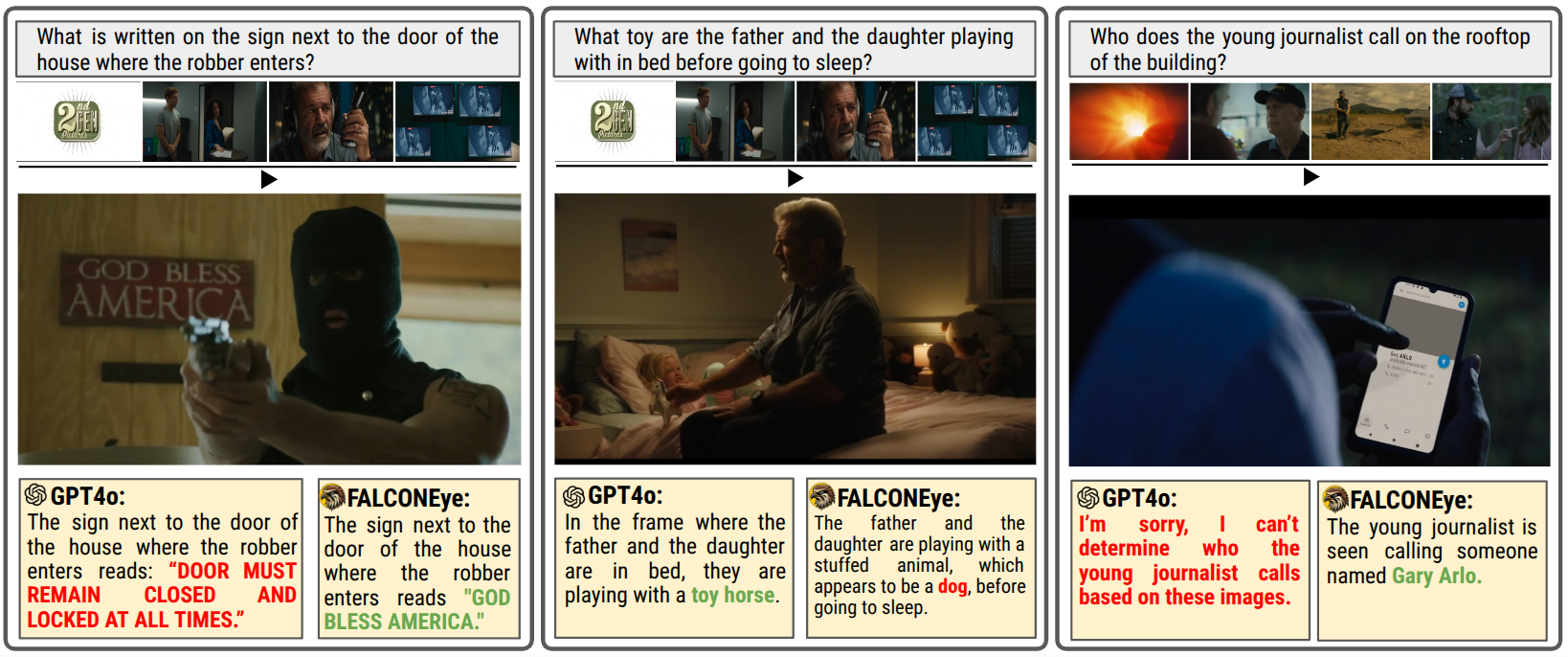}
    \caption{Answer comparison between GPT4o and FALCONEye for three example questions, showing the frame that contain the answer.}
    \label{fig:gpt4o_vs_falconeye}
\end{figure*}

%Table~\ref{tab:calibexp} presents the calibration analysis for LLaVA-Video, LLaVA-OV, and Qwen2.5-VL. Surprisingly, all methods exhibit strong calibration in MCQs (Fig.~\ref{fig:calibrationmodels} of Supp. Material), with LLaVA-Video achieving the highest performance. For OQs, Qwen2.5-VL significantly outperforms other approaches, particularly in CC, as it maintains a high percentage of confident predictions above the threshold (0.8) while keeping calibration error low. The calibration plots for Qwen2.5-VL are shown in Fig.~\ref{fig:qwen25_vl_calibration} and the others are in the supplementary material (Fig.~\ref{fig:calibrationmodels}).

\paragraph{Models-.} Regarding model selection, we compare reliability diagrams for LLaVA-Video, LLaVA-OneVision, and Qwen2.5-VL, evaluated on both MCQs and OQs. For MCQs, all models show strong calibration, with low calibration error, specially for high confidence values (Fig.~\ref{fig:calibrationmodels} (g-l)). However, for OQs, the distribution of high-confidence answers differs significantly. Both LLaVA-Video and LLaVA-OneVision have a much smaller proportion of answers with 0.9 confidence, whereas Qwen2.5-VL not only produces a higher number of high-confidence answers but also demonstrates extremely low calibration error for those predictions.

% \begin{figure*}[!htb]
%     \centering
%     \includegraphics[width=0.8\linewidth]{images/gpt4o_vs_falconeye.png}
%     \caption{Answer comparison between GPT4o and FALCONEye for three example questions, showing the frame that contain the answer.}
%     \label{fig:gpt4o_vs_falconeye}
% \end{figure*}

\subsection{FALCONEye vs GPT4o}
Figure \ref{fig:gpt4o_vs_falconeye} shows three FALCON-Bench example questions and compare the responses from GPT4o and FALCONEye.

\subsection{FALCON-Bench vs QAEgo4D}
The QAEgo4D \cite{Baermann_2022_qaego4d} dataset is the closest to FALCON-Bench in terms of the task definition as it also addresses the problem of VAS, that the authors call \emph{question answering visual language grounding} (VLG). They also have open ended questions. However, besides the time differences in the video durations --FALCON-Bench with an average duration of $\sim 80$ minutes while QAEgo4D has an average duration of $\sim 8$ minutes--, the type of questions are completely different. QAEgo4D questions are automatically generated from the sparse video narration, focusing on the main object or action: \emph{Q: What did I Put in the Pan?}--\emph{A: cheese}, or \emph{Q: What paint can did I open?}--\emph{A:black paint}. Meanwhile, FALCON-Bench questions are human curated to be challenging: \emph{Q: What is the number of the train that crosses paths with Lightning McQueen at night?}--\emph{A:A113}; \emph{Q:What message about the flu appears on a city building?}--\emph{A:Bovril nourishes you to resist 'flu} or \emph{Q:Which player of Chelsea received a red card?}--\emph{A: Thibaut Courtois}.

%Besides, at the moment of the submission, the Ego4D NLQ dataset, which is used in QAEgo4D has a bug where 14\% of queries have a near-0 query window, artificially reporting a lower performance of all methods.

\end{document}